%% file: main.tex
\documentclass[10pt,twocolumn,letterpaper]{article}

\usepackage{cvpr}              

\usepackage{booktabs}
\usepackage{adjustbox}
\usepackage{amssymb}
\usepackage{pifont}
\usepackage{subcaption}
\usepackage{wrapfig}

\definecolor{cvprblue}{rgb}{0.21,0.49,0.74}
\usepackage[pagebackref,breaklinks,colorlinks,allcolors=cvprblue]{hyperref}

\newcommand{\cmark}{\ding{51}}%
\newcommand{\xmark}{\ding{55}}%

\def\paperID{214} 
\def\confName{CVPR}
\def\confYear{2025}

\title{Compass Control: Multi Object Orientation Control for Text-to-Image Generation}

\author{
  \begin{tabular}[t]{c}
    Rishubh Parihar$^{1}$\footnotemark[1] \quad
    Vaibhav Agrawal$^{2*}$\footnotemark[2]  \quad
    Sachidanand VS$^{1}$ \quad 
    R. Venkatesh Babu$^1$\\ 
    \end{tabular}%
  \quad
  \and
  \begin{tabular}[t]{c} 
    $^1$IISc Bangalore \quad 
    $^2$IIIT Hyderabad\\
  \end{tabular}%
}

\begin{document}
\twocolumn[{
    \begin{@twocolumnfalse}  
        \maketitle
        \thispagestyle{plain}  
        \vspace{-5mm}  
        \begin{center}
            \includegraphics[width=\textwidth]{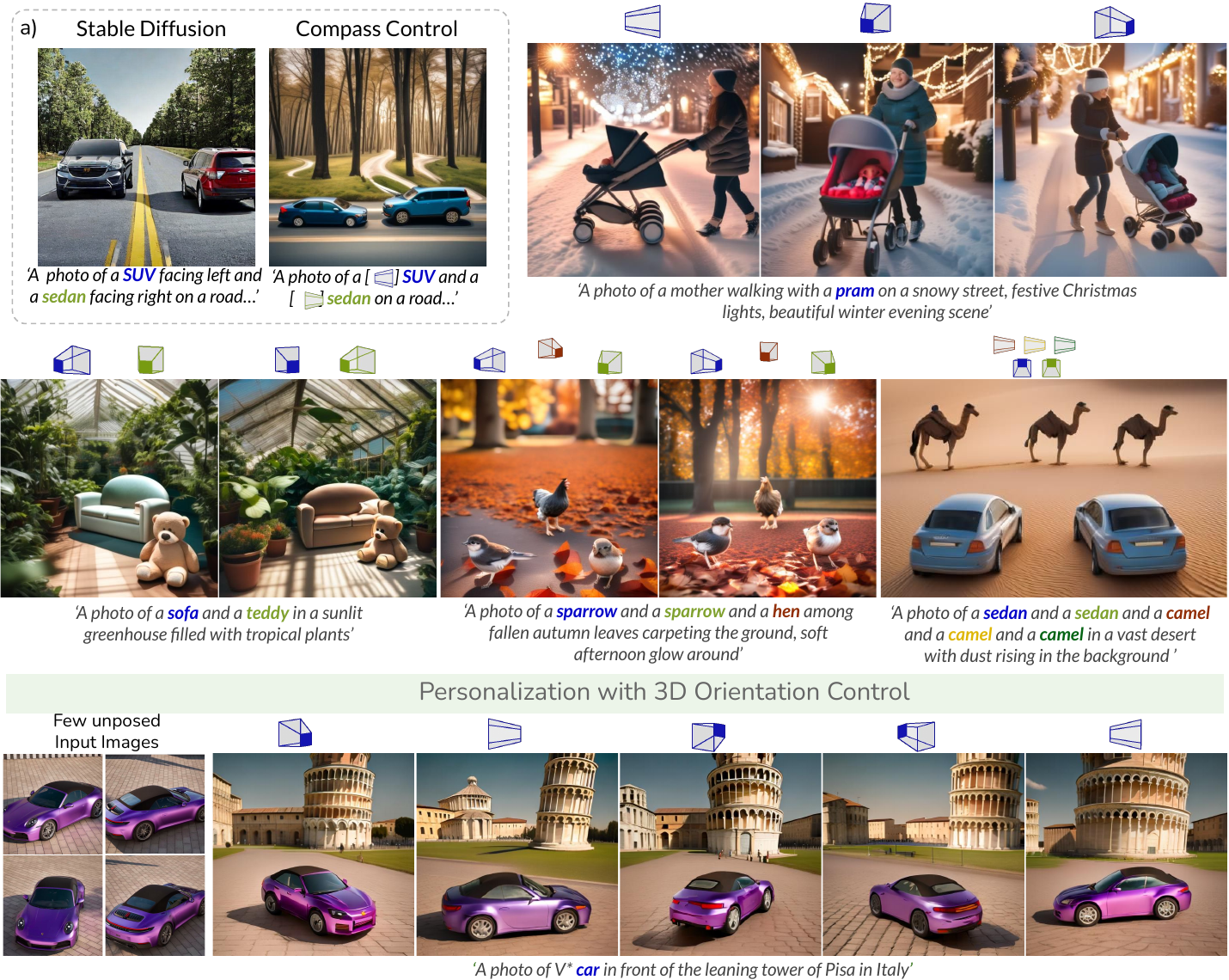}
            \vspace{-2mm}
            \captionsetup{type=figure}
            \captionof{figure}{
                We present \textit{Compass Control}, a method to generate multi-object scenes with orientation control from text-to-image diffusion models. Given a text \textit{prompt} and an orientation of each object (shown as \textcolor{blue}{frustum}, the colored face is the forward direction), our method generates scenes that align with both the prompt and specified orientations. Additionally, with a few ($\approx10$) unposed images of a new object, our model is personalized to generate the object in target orientations.
            }
            \label{fig:teaser}
        \end{center}

        \vspace{5mm}  
    \end{@twocolumnfalse}
}]

\renewcommand{\thefootnote}{\fnsymbol{footnote}} 
\footnotetext[1]{equal contribution.}
\footnotetext[2]{work done during an internship at VAL, IISc}

\input{sec/0_abstract}   

\let\origaddcontentsline\addcontentsline
\renewcommand{\addcontentsline}[3]{} 
\input{sec/1_intro}
\input{sec/2_relatedwork}
\input{sec/3_method}

\input{sec/4_experiments}
\input{sec/5_conclusion}

{
    \small
    \bibliographystyle{ieeenat_fullname}
    \bibliography{main}
}

\let\addcontentsline\origaddcontentsline
\input{sec/X_suppl}

\end{document}

%% file: sec/0_abstract.tex
\begin{abstract} 
Existing approaches for controlling text-to-image diffusion models, while powerful, do not allow for explicit 3D object-centric control, such as precise control of object orientation. In this work, we address the problem of multi-object orientation control in text-to-image diffusion models. This enables the generation of diverse multi-object scenes with precise orientation control for each object. The key idea is to condition the diffusion model with a set of orientation-aware \textbf{compass} tokens, one for each object, along with text tokens. A light-weight encoder network predicts these compass tokens taking object orientation as the input. The model is trained on a synthetic dataset of procedurally generated scenes, each containing one or two 3D assets on a plain background. However, direct training this framework results in poor orientation control as well as leads to entanglement among objects. To mitigate this, we intervene in the generation process and constrain the cross-attention maps of each compass token to its corresponding object regions. The trained model is able to achieve precise orientation control for a) complex objects not seen during training and b) multi-object scenes with more than two objects, indicating strong generalization capabilities. Further, when combined with personalization methods, our method precisely controls the orientation of the new object in diverse contexts. Our method achieves state-of-the-art orientation control and text alignment, quantified with extensive evaluations and a user study. \href{rishubhpar.github.io/compasscontrol.github.io}{project page}

\vspace{-4mm}

\end{abstract}

%% file: sec/1_intro.tex
\vspace{-2mm}
\section{Introduction}
\label{sec:intro} 

\vspace{-2mm}
\noindent 
Imagine a visual artist aiming to create a scene featuring two cars facing each other. They prompt a text-to-image model with - \textit{`A photo of a sedan facing right and an SUV facing left'}. However, as shown in Fig.~\ref{fig:teaser} (a), the generated image may not always accurately capture the intended object orientations. Moreover, relying on text prompts to control object orientation (e.g., \textit{`facing right'}) is imprecise and requires iterative prompting. \textit{Can we design an alternate interface for text-to-image models that accepts target orientation angle as input along with the text prompts?} Such an interface will allow for precise orientation control for each object, eliminating the need for iterative prompt adjustments and streamlining the creative process.



Several works have been proposed to achieve finer control in text-to-image models such as changing object appearance, scene layouts or image style ~\cite{controlnet, sketch-guided, ye2023ipadapter, gengmotion, song2023objectstitch, chen2024anydoor, cao2023masactrl, hertz2022prompt, brack2024ledits++, patashnik2023localizing}. While effective for controlling 2D attributes of the image, these methods fail to accurately control 3D attributes. More recently, several methods have been proposed to control 3D properties in text-to-image models, such as camera viewpoint~\cite{cd-360, view-neti}, scene lighting~\cite{cont-words}, or scene layout using 3D bounding boxes~\cite{loosecontrol}. However, these approaches either require dense 3D information, like multi-view images or accurate 3D bounding boxes, or limited to simple single-object scenes (Tab.~\ref{tab:charac}). In this work, we present a novel interface to condition text-to-image diffusion models to generate multi-object scenes with precise 3D orientation control, without the need for multi-view images or 3D bounding boxes.

Text-to-image (T2I) diffusion models enable the generation of objects with specific attributes via text prompts (e.g., \textit{`a red car'}). Motivated by this, we encode the object orientation as an additional attribute in the text embedding space of the T2I model. 
Specifically, we introduce a special token, dubbed as \textit{\textbf{compass}} token ($\mathbf{c}$) along with each token in the prompt ({e.g., \textit{`A photo of $\mathbf{c_1}$ SUV and $\mathbf{c_2}$ sedan on a road.'}). Each \textit{compass} token is predicted by a lightweight encoder model taking the prescribed orientation angle as input. This formulation preserves the original interface of the base T2I model and enables precise object-centric orientation control in multi-object scenes. We train the encoder model and fine-tune the denoising U-Net with LoRA~\cite{hu2021lora} on a synthetic dataset of scenes containing one or two 3D assets placed in diverse layouts on an empty floor.

We discover that directly injecting \textit{compass} tokens with the prompt tokens leads to poor orientation control, as the \textit{compass} token attends to irrelevant image regions, limiting its influence on its corresponding object (Fig.~\ref{fig:attn-reg-method-ablate}(a)).  To address this, we propose \textbf{C}oupled \textbf{A}ttention \textbf{L}oca\textbf{l}ization (\textbf{CALL}) mechanism, where we constrain the cross-attention maps of the \textit{compass} token and its corresponding object token within a 2D bounding box. This results in a tight \textit{coupling} between the two tokens, enabling the \textit{compass} token to precisely control the orientation for its corresponding object. Additionally, for multi-object scenes, this results in an appropriate binding between each \textit{compass} token and its corresponding object token, leading to disentangled orientation control of individual objects (Fig.~\ref{fig:attn-reg-method-ablate}(b)). 

\begin{table}[]
\caption{\textbf{Comparison with Related works.} Our approach uniquely allows for object-centric orientation control and generalize to novel categories without any explicit 3D representation.}
\vspace{-2mm}
\begin{adjustbox}{width=\linewidth}
\begin{tabular}{@{}c|ccccc@{}}
\toprule
Method                                                       & CD-360~\cite{cd-360}                          & LooseControl~\cite{loosecontrol}                        & Cont-3D-Words~\cite{cont-words}                   & ViewNeTI~\cite{view-neti}                        & Ours                            \\ \midrule
Input                              & Cam Pose & 3D boxes & Cam Pose & Cam Pose & Orientation \\
3D conditioning                                                   & Explicit                        & Explicit                            & Implicit                        & Implicit                        & Implicit                        \\
Novel classes                                                     & \xmark                              & \cmark                                 & \xmark                             & \cmark                             & \cmark                             \\
\begin{tabular}[c]{@{}c@{}}Multiple Object\\ Control\end{tabular} & \cmark                             & \cmark                                 & \xmark                              & \xmark                              & \cmark                             \\ \bottomrule
\end{tabular}
\end{adjustbox}
\label{tab:charac}
\vspace{-6mm} 
\end{table}


The proposed approach achieves precise orientation control for unseen objects (e.g., pram) and can generalize to scenes with more than two objects, despite being trained on one and two object scenes only (see Fig.~\ref{fig:teaser}). Further, given a few unposed images of a real object, we can personalize the model to control the orientation of the new object. We evaluate our method against several baselines, achieving superior performance both quantitatively and in a user study. In summary, our primary contributions are:



\begin{enumerate}
    \item \textit{Compass Control} - A method for conditioning text-to-image diffusion models on object orientation, enabling precise orientation control for individual objects in multi-object scene generation. 
    
    \item \textit{Coupled Attention Localization} - A mechanism to restrict the influence of the input object orientation
    to the corresponding object, ensuring effective object-centric orientation control and object disentanglement.
    
    \item Strong generalization of \textit{Compass Control} for precise orientation control to \textit{unseen} objects and complex multi-object scenes, though trained on simple synthetic scenes. 
    
    \item \textit{Personalization of Compass Control} - Given a few unposed images of a real object, our method can perform orientation control of the new object in diverse contexts. 
    
\end{enumerate}

%% file: sec/2_relatedwork.tex
\vspace{-2mm}
\section{Related work}
\label{sec:related}

\vspace{-2mm}
\noindent 
\textbf{Controlled generation in T2I models.} Several works have been proposed to achieve fine-grained control in text-to-image diffusion models~\cite{rombach2022high,dalle2,imagen}. Recent works resort to manipulation of the text embeddings ~\cite{ge2023expressive-t2i, kawar2023imagic, zhang2023prospect, voynov2023p+, parihar2024precisecontrol}, or attention maps in the diffusion U-Net~\cite{cao2023masactrl,hertz2022prompt,diffusion-handles,bar2023multidiffusion,dahary2024be_yourself, avrahami2024diffuhaul, patashnik2023localizing, sketch-guided, attend-excite, kim2023dense, voynov2023sketch, rangwani2024crafting} for controlling the generated image. Additional encoder models can be trained to condition the  T2I models on a new modality such as depth, bounding boxes, or object identity~\cite{controlnet,ye2023ipadapter,li2023gligen}. Another set of works personalizes the diffusion model given with a few subject images~\cite{ruiz2023dreambooth,kumari2023multi} enabling generation of the learned subject in different backgrounds. Recent work on guiding diffusion models ~\cite{epstein2023selfguidance,luo2024readout,parihar2024balancing} allows inference time control over the scene contents, allowing the control of object location, appearance, shape, and skeleton pose. However, these controls are limited to 2D.

\vspace{1mm} 
\noindent 
\textbf{3D-aware image editing.} 
Recent works leverage the rich generation capability of the T2I model to perform 3D aware editing~\cite{wang2024diffusioncritc,diffusion-handles,sajnani2024geodiffuser}. Specifically, they use the input scene depth as an additional input and use it to warp the internal features of the diffusion models. This enables zero-shot geometric 3D edits such as translating or rotating an object. However, these methods are limited to the editing of a single object. Another line of work uses multi-view input images and trains an implicit 3D representation such as a radiance field in the diffusion feature space~\cite{patashnik2024consolidating,cd-360} to perform 3D consistent editing. More recently, few works leverage 3D Gaussian splat representation along with T2I models to perform scene editing ~\cite{chen2023gaussianeditor,luo20243d}. However, the above methods require scene-specific training and require multi-view images of accurate depth maps as input. Another direction explores large-scale training of diffusion model on a specific dataset~\cite{neural-assets,object-3dedit,jabri2023dorsal} allowing for 3D scene editing. However, these methods fail to generalize to in-the-wild real-world scenes outside the distribution of training datasets.

\noindent 
\textbf{3D control in generation.} Earlier works train generative models from scratch with explicit scene controls such as 3D blobs~\cite{wang2023blobgan}, or radiance fields of individual objects~\cite{nguyen2019hologan, niemeyer2021giraffe, xue2022giraffehd, blendnerf, bautista2022gaudi, Kathare_2025_WACV} to control the generated scene. Recent works have shown the existence of 3D properties in text-to-image diffusion models~\cite{diffusion3d-understand1,diffusion3d-understand2,diffusion3d-understand3,view-neti,dhiman2024reflecting}. This has fuelled the research that leverages this knowledge from T2I models for 3D generation. A set of works lifts this knowledge to 3D by distilling from pre-trained T2I models~\cite{poole2022dreamfusion,wang2023luciddreaming,zhou2025dreamscene360,vilesov2023cg3d,lin2023magic3d,yu2024wonderworld}. 
Another set of recent works~\cite{view-neti, loosecontrol,cont-words} leverage this underlying 3D knowledge for controlling 3D properties of the generated scene with an additional conditioning mechanism. ViewNeTI~\cite{view-neti} learns a 3D view token to control the camera view for the task of novel-view synthesis from a dataset of multi-view images. ViewNeTI is natively designed for novel view synthesis but is unable to generate objects in diverse contexts. A continuous word representation is trained in~\cite{cont-words} for 3D scene properties such as orientation and lighting on a single 3D mesh. However, both these approaches are limited to simple scenes and control only the global view angle. The closest to our work is~\cite{loosecontrol} that conditions the T2I model on loose depth maps using ControlNet~\cite{controlnet}. Loose depth maps are created using 3D object boxes and scene boundaries to enable 3D object-centric control. However, this approach relies on precise 3D boxes, making it cumbersome at inference. In contrast, our method only requires coarse 2D bounding boxes and orientation angles, offering a more user-friendly solution.

\begin{figure}[t]
    \centering
    \includegraphics[width=\linewidth]{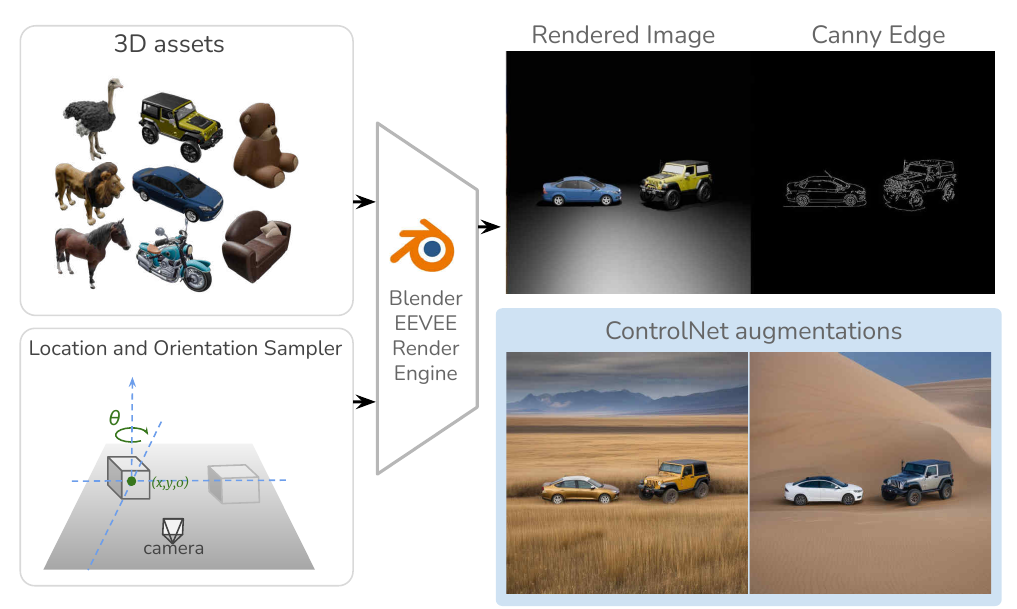}
    \vspace{-6mm}
    \caption{\textbf{Synthetic data generation.} We curate 10 diverse 3D assets, and render them in diverse layouts and orientations in Blender~\cite{blender}. The rendered scenes are augmented with realistic generations from Canny~\cite{canny1986computational} ControlNet~\cite{controlnet}. The final dataset consists of one and two object scenes.}
    \label{fig:dataset-gen}
    \vspace{-4mm}
\end{figure}  

%% file: sec/3_method.tex
\vspace{-2mm}
\section{Method}
\label{subsec:method}
\vspace{-3mm}

\vspace{1mm}
\noindent 
\textbf{Text-to-Image Diffusion Models.} Diffusion models, when directly applied in the pixel space, are computationally expensive due to their iterative nature. To mitigate this, latent diffusion models~\cite{rombach2022high} apply the diffusion process in the smaller resolution latent space of a pretrained autoencoder. Further, the generation can be conditioned on text by injecting text features into the diffusion U-Net with additional cross-attention layers. The cross-attention maps allow for precise control during generation ~\cite{hertz2022prompt2prompt,epstein2023selfguidance}.

\subsection{Dataset}
\label{subsec:dataset} 
\vspace{-1mm} 

 \textbf{Synthetic scene generation:} We curated a list of 10 diverse 3D assets from the web: ostrich, helicopter, shoe, jeep, teddy bear, lion, sedan, horse, motorbike and sofa to aid in model generalization (see Fig.~\ref{fig:dataset-gen}). Our dataset consists of 1000 one-object and $7900$ two-object scenes, rendered in Blender~\cite{blender}. For each image, we save the 2D bounding boxes and the orientations of the objects. The objects are placed at varied locations and in varied orientations to increase layout diversity.

\vspace{1mm} 
\noindent 

\noindent 
\textbf{Augmentations:} Directly training on this dataset results in overfitting to the plain background and the black floor, as shown in ablations. To address this, we augment the dataset using ControlNet~\cite{controlnet} to place objects in diverse backgrounds while retaining known orientations. For each rendered image, we extract its Canny~\cite{canny1986computational} edge map to condition ControlNet to generate the objects in diverse contexts (for e.g., `\textit{in a garden...}', `\textit{near a lake...}', etc). This approach preserves object orientations while altering their appearance. We manually filter out the inconsistent augmented images. Further details on the dataset creation process can be found in the Suppl. Sec.\textcolor{cvprblue}{H}.

\vspace{1mm}
\noindent 
\textbf{Orientation convention:} In this paper, we parameterize orientation with a single angle $\theta$, rotation around the up axis in the world coordinate system (pointing towards the sky). We define $\theta=0$ as a reference when the object faces exactly towards the right (e.g., sedan in Fig.~\ref{fig:dataset-gen}). We parameterize with a single orientation angle, as most of our objects are land objects, and only this rotation axis results in plausible object orientations. However, our method is not limited to a single orientation angle, and we present results in Suppl. Sec.\textcolor{cvprblue}{B} for conditioning on three orientation angles. 

\begin{figure}[t]
    \centering
    \includegraphics[width=0.90\linewidth]{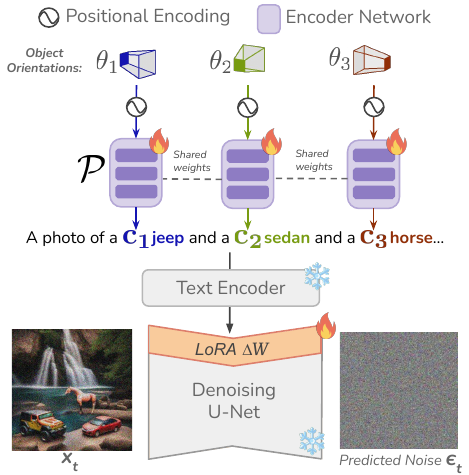}
    \vspace{-3mm}
    \caption{\textbf{\textit{Compass Control}.} Given an orientation angle $\theta_j$, we project it to a \textit{compass} token with a lightweight encoder model. The \textit{compass} tokens are interleaved with the text tokens (as shown in the figure) and passed through the text encoder. The outputs of the text encoder are used to condition the denoising process in the U-Net. We train $\mathcal{P}$ and also fine-tune the U-Net using LoRA~\cite{hu2021lora}.}
    \label{fig:method-figure}
    \vspace{-6mm}
\end{figure} 

\subsection{Compass Control}
\label{subsec:method}
\vspace{-1mm}

Given a text prompt $\mathcal{T}$ consisting of $N$ object names $\{o_1, o_2, ..., o_N\}$ (e.g., \textit{`A photo of a \textbf{jeep} and a \textbf{sedan} and a \textbf{horse} in a garden'}) and their corresponding 3D orientation angles $\{\theta_1, \theta_2, ..., \theta_N\}$ we introduce a set of \textit{compass} tokens $\{ \mathbf{c_1}, \mathbf{c_2}, ..., \mathbf{c_N} \}$ to control the orientations of the respective objects. A \textit{compass} token $\mathbf{c_n}$ is an embedding in the input space of the text encoder. It is predicted by a lightweight MLP encoder network $\mathcal{P}$ (see Fig.~\ref{fig:method-figure}), which takes as input the object orientation angle $\theta_k$. The compass tokens are prepended before their corresponding object tokens (e.g., \textit{`A photo of a $\mathbf{c_1}$ \textbf{jeep} and a $\mathbf{c_2}$ \textbf{sedan} and a $\mathbf{c_3}$ \textbf{horse} in a garden'}), and passed through the text encoder. The outputs of the text encoder are used to condition the denoising U-Net. 

\begin{figure}[t]
    \centering
    \includegraphics[width=0.95\linewidth]{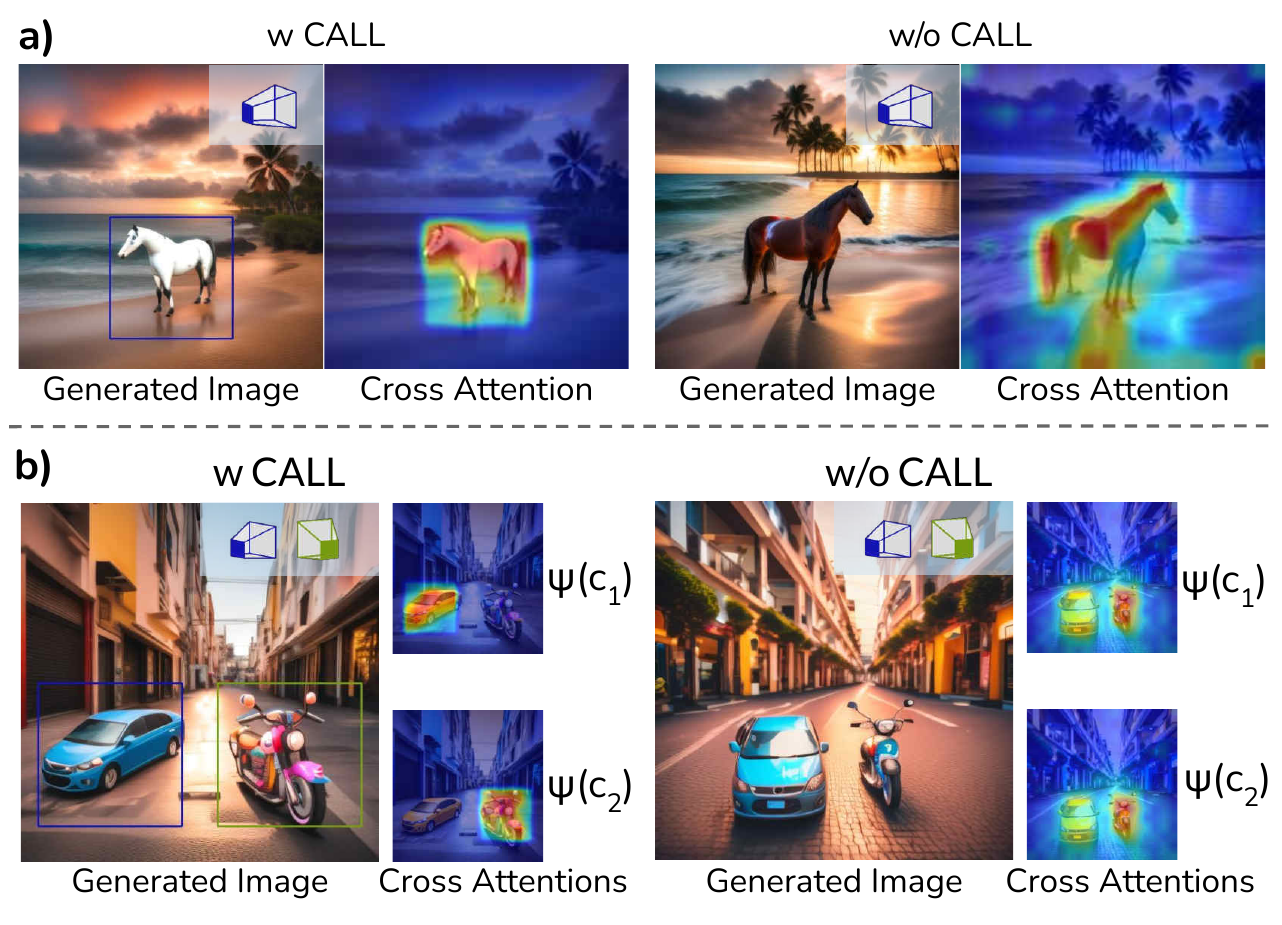}
    \vspace{-3mm}
    \caption{\textbf{Binding the \textit{compass} tokens: } We visualize the averaged cross attention of the \textit{compass} token(s) when training with CALL (shown on the left) and without it (shown on the right). CALL localizes the influence of the \textit{compass} token at the \textit{correct} regions, which (a) improves orientation control (b) disentangles orientations in multi-object scenes. In (b), $\mathbf{c_1}$ and $\mathbf{c_2}$ are compass tokens for car and motorbike, respectively.} 
    \label{fig:attn-reg-method-ablate} 
\end{figure}

However, directly training the above framework on the dataset from Sec.~\ref{subsec:dataset} fails to learn accurate orientation control. We hypothesize that this is because the added \textit{compass} tokens are unrestricted and can attend to irrelevant image regions, limiting their influence on their  corresponding objects. This is evident in the cross-attention maps for the \textit{compass} tokens, which are indeed diffused in other image regions (see Fig.~\ref{fig:attn-reg-method-ablate} (a)). Furthermore, this becomes a severe issue in multi-object scenes, where a single \textit{compass} token can attend to multiple objects, resulting in the entanglement between different \textit{compass} tokens(see Fig.~\ref{fig:attn-reg-method-ablate} (b)). Existing works have shown that cross-attention maps closely control the image layouts~\cite{hertz2022prompt2prompt} in the generated image. Motivated by this, we design a cross-attention localization approach.

\subsection{Coupled Attention Localization (CALL)} 
\vspace{-1mm}
\label{subsec:attn-reg}
Our key idea is to constrain the cross-attention maps for both the \textit{compass} token and the corresponding object token inside a given 2D bounding box. This enables tight association between the object and the \textit{compass} tokens. Additionally, it enables explicit control over the object location during generation. Specifically, during training, we use the saved object bounding boxes $\{b_1, ... b_N\}$ to compute a set of \textit{loose square} bounding boxes $\{b^l_1, ... b^l_N\}$. The side length $a_n$ of the loose box $b^l_n$ is computed as $a_n = \lambda * max(h_{b_n},w_{b_n})$, where $h_{b_n}$ and $w_{b_n}$ are the height and width of the object box $b_n$ and $\lambda>1$ is a padding factor controlling the \textit{looseness} of the box. Next, we compute a binary mask $m_n$ from the loose bounding box $b^l_n$, such that m is $0$ inside the box $b^l_n$ and $-\infty$ outside. We use it to mask out the cross attentions ($\Psi$) of the object token $o_n$ and \textit{compass} token $\mathbf{c_n}$ as follows:

\vspace{-3mm}
\begin{equation*}
\small
    \Psi(\mathbf{c_n}) = \text{softmax}\Biggl( m + \frac{Q{K(\mathbf{c_n})}^T}{\sqrt{d_K}}\Biggr)
\end{equation*}
\vspace{-2mm}
\begin{equation*}
\small
    \Psi(\mathbf{o_n}) = \text{softmax}\Biggl( m + \frac{Q{K(o_n)}^T}{\sqrt{d_K}} \Biggr)
\end{equation*}
\noindent 
where the query feature $Q$ comes from the U-Net features, and the key features $K(\mathbf{c_n})$ and $K(\mathbf{o_n})$ come from the respective tokens. This masking operation is performed at all the diffusion timesteps and cross-attention layers. We find that using a loose mask is highly effective, providing greater flexibility during the generation process. This attention localization mechanism is dubbed as \textit{Coupled Attention Localization}, or \textit{CALL} for short. Adding CALL mechanism during training and inference has key advantages for composing multi-object scenes: 
\textit{a)} Appropriate binding of each compass token $\mathbf{c_n}$ with its corresponding object token $o_n$ leads to disentangled orientation control of individual objects.
\textit{b)} Constraining the cross-attention for the object tokens $o_n$ to \textit{non-overlapping} bounding boxes results in disentanglement between the objects themselves (a known issue in T2I models~\cite{attend-excite}), \textit{enabling strong generalization to complex scenes with multiple objects}.

\begin{figure}[t]
    \centering
    \includegraphics[width=\linewidth]{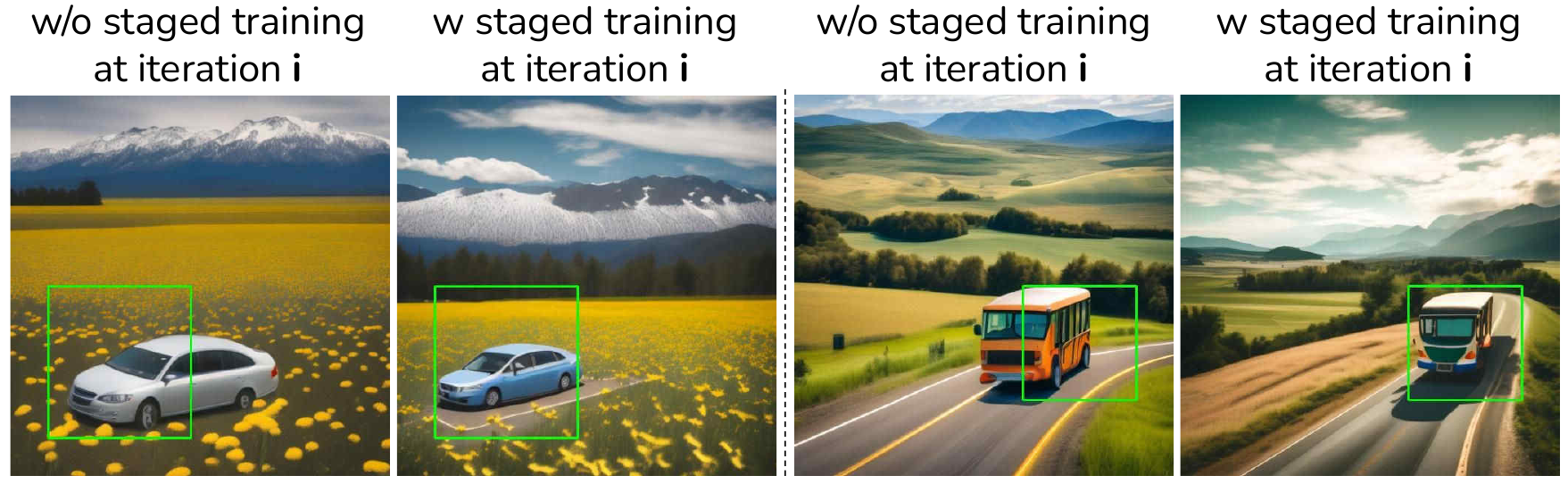}
    \vspace{-4mm}
    \caption{\textbf{Staged training} results in improved adherence of objects to the bounding boxes, leading to orientation learning.}
    \label{fig:stage-training-ablate}
    \vspace{-4mm}
\end{figure}

\vspace{1mm}
\noindent 
\textbf{Training.} We train our encoder model $\mathcal{P}$ and fine-tune the denoising U-Net with LoRA~\cite{hu2021lora} on the synthetic dataset from Sec.~\ref{subsec:dataset}. The LoRA training is extremely parameter efficient and preserves the behavior of the base T2I model. We use the proposed CALL mechanism for effective learning of orientation control. However, for the effective working of CALL, the object must be generated within the loose bounding box, as the \textit{compass} token's influence is restricted to this region. To this end, we first train on simple single-object scenes, to first learn the bounding box adherence for the generated object, and then train on a mix of both single and two-object scenes thereafter. We contrast this two-staged training procedure with the single-stage training at an intermediate training iteration in Fig.~\ref{fig:stage-training-ablate}. The generated objects adhere to the bounding box better in the two-staged training compared to single-stage training This leads to effective learning of orientation control, as we have shown in ablative experiments (Sec.~\ref{subsec:experiments}).  

\vspace{1mm}
\noindent 
\textbf{Inference.} During the inference phase, \textit{Compass Control} expects the text prompt containing the objects, desired orientations, and optional coarse 2D bounding boxes as input. \textit{Using loose bounding boxes during training offers a significant advantage here, as we can even spawn non-overlapping boxes heuristically}, as shown in the Suppl. Sec.\textcolor{cvprblue}{F}. We use the text tokens and the \textit{compass} tokens together to condition the diffusion model.

\noindent 
\vspace{-4mm}
\subsection{Personalization}
\label{subsec:personlize} 
\vspace{-1mm}

The design of \textit{Compass Control} as a conditioning mechanism preserves the original capabilities of T2I models, such as personalization. Given a few unposed images of an object ($\approx 10$), we apply Dreambooth~\cite{ruiz2023dreambooth} with LoRA and associate a special token $\hat{u}$ for the input object using \textit{Compass Control}'s fine-tuned UNet. During inference, we can generate the object in desired orientation $\theta$ with simple prompts; e.g., \textit{`A photo of a $\mathbf{c(\theta)}$ $\hat{u}$ car on the beach.'}, where $\mathbf{c(\theta)}$ is the \textit{compass} token. 

%% file: sec/4_experiments.tex
\vspace{-2mm} 
\section{Experiments}
\label{subsec:experiments}
\begin{figure*}[t]
    \centering
    \includegraphics[width=\linewidth]{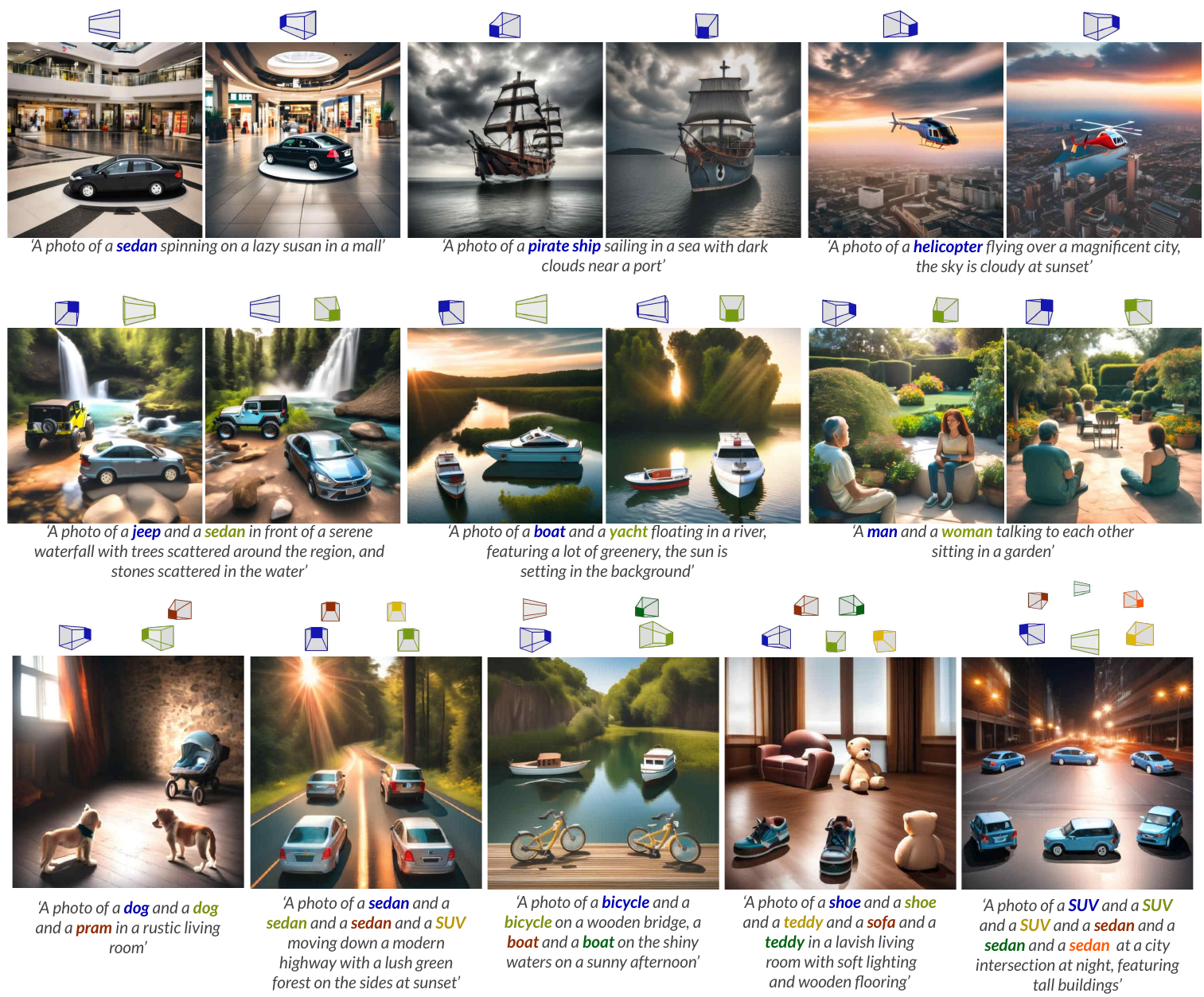}
    \vspace{-3mm}
    \caption{\textbf{Main Results.} \textit{Compass Control} generates complex scenes aligned with the text prompts and the orientations (shown as \textcolor{blue}{frustum}, the colored face is forward direction). It generalizes well to unseen object categories - \textit{pirate ship, boat, yacht, bicycle, pram, dog, and even human}. Further, it generates high-quality compositions of several objects, despite being \textit{trained only on one and two object scenes.}}
    \label{fig:qual-ours}
    \vspace{-4mm}
\end{figure*}

\vspace{-1mm}
\subsection{Experiment setup}
\label{subsec:exp-overview} 
\vspace{-1mm} 

\textbf{Dataset.} We use the synthetic dataset from Sec.~\ref{subsec:dataset} consisting of $8900$ rendered images and $6010$ augmented images consisting of one and two object scenes for training our model. For quantitative evaluation, we construct a test set of $10$ scene prompts having one/two object names generated from ChatGPT~\cite{chatgpt}. We use a mix of seen objects (\textit{horse, jeep, sedan, sofa, teddy, lion}) and unseen objects (\textit{boat, dolphin, ship, SUV, tractor}) in the prompts. For each object and prompt combination, we use a set of $10$ randomly sampled orientations. The list of input text prompts and orientations is given in the Supp. Sec.\textcolor{cvprblue}{L}. 

\vspace{1mm}
\noindent
\textbf{Implementation Details.} We use Stable Diffusion v2.1~\cite{rombach2022high} as our base T2I model and use LoRA rank $4$ for fine-tuning its UNet. We present additional results on Stable Diffusion-XL~\cite{podell2023sdxlimprovinglatentdiffusion} in Suppl. Sec. \textcolor{cvprblue}{D}. Our encoder model $\mathcal{P}$ is a lightweight MLP: three linear layers with ReLU. We train our model for $25,000$ steps with a batch size of $4$ with AdamW optimizer and a fixed learning rate of $10^{-4}$. We keep the bounding box padding $\lambda=1.2$ for CALL. The training takes 24 hours on a single A6000 GPU.

\vspace{1mm}
\noindent 
\textbf{Evaluation Metrics.} We evaluate for a) \textit{Text Alignment} using CLIP similarly; b) \textit{\%Object Generation} - we evaluate the presence of the intended object using Grounding-DINO~\cite{liu2023grounding} and threshold on the objectness score for each object in the prompt. c) \textit{Angular Error} - to evaluate the orientation consistency, we compute the Angular error (in radians) between the input orientation angle $\theta$ and the orientation of the generated object using a pretrained orientation predictor. Further details about the orientation predictor and the implementation details of the metrics are in the Suppl Sec.\textcolor{cvprblue}{I}.


\vspace{1mm}
\noindent 
\textbf{Baselines.} As no prior method tackles our same task, we compare against methods that allow for either camera pose control or 3D object pose control in text-to-image models: a) Continuous 3D Words (\textit{Cont-3D-Words}) ~\cite{cont-words}: Following their exact setup, we train a 3D word for controlling the object orientation on renderings of a single 3D asset - Sedan and its ControlNet augmentations. b) \textit{ViewNeTI}~\cite{view-neti}: In contrast to Cont-3D-Words, ViewNeTI allows for training on multiple 3D assets; for fair evaluation, we train ViewNeTI on our training dataset and condition the T2I on object orientation instead of 3D camera pose. Notably, both of these methods are limited to a global view control. Hence, we evaluate on only single-object scenes. c) \textit{LooseControl}~\cite{loosecontrol}, allows for multi-object control by conditioning on loose depth maps (Fig.~\ref{fig:baseline-compare}) formed by 3D object boxes. We use template 3D bounding boxes for each test object and place them in the scene, with random orientation and location (similar to Sec.~\ref{subsec:dataset}). For a fair comparison, we use the 2D boxes corresponding to the 3D boxes in our outputs. Notably, LooseControl does not take exact orientation as input as a 180 flipped box also has the same \textit{loose} depth; we consider this while computing the Angular error. Further, it requires the user to provide \textit{accurate 3D bounding boxes during inference,} which is cumbersome, whereas our method requires only loose 2D boxes. Additional baseline details are in Suppl. Sec.\textcolor{cvprblue}{J}.

\subsection{Main Results}
\vspace{-1mm}
\textbf{Qualitative results.} 
We present our method's results in Fig.~\ref{fig:qual-ours}. Our method is able to generate complex multi-object scenes with precise orientation control of individual objects, even though it is \textit{trained with single and two object scenes.} Further, it is able to generalize well to challenging unseen objects such as \textit{humans} and \textit{prams}. Interestingly, there was no water-based subject in the training dataset, yet our method can achieve precise orientation control for a \textit{ship}, \textit{yacht}, and \textit{boat}. These strong generalization capabilities of \textit{Compass Control} can be attributed to effective attention constraining with CALL and diversity in the 3D assets used for training. The conditioning mechanism of Compass Control is generalized, and we present results for jointly controlling all \textit{three orientation angles}, \textit{camera elevation}, and \textit{object scale} in Suppl. Sec.\textcolor{cvprblue}{B} \& \textcolor{cvprblue}{C}.

\begin{figure}[t]
    \centering
    \includegraphics[width=\linewidth]{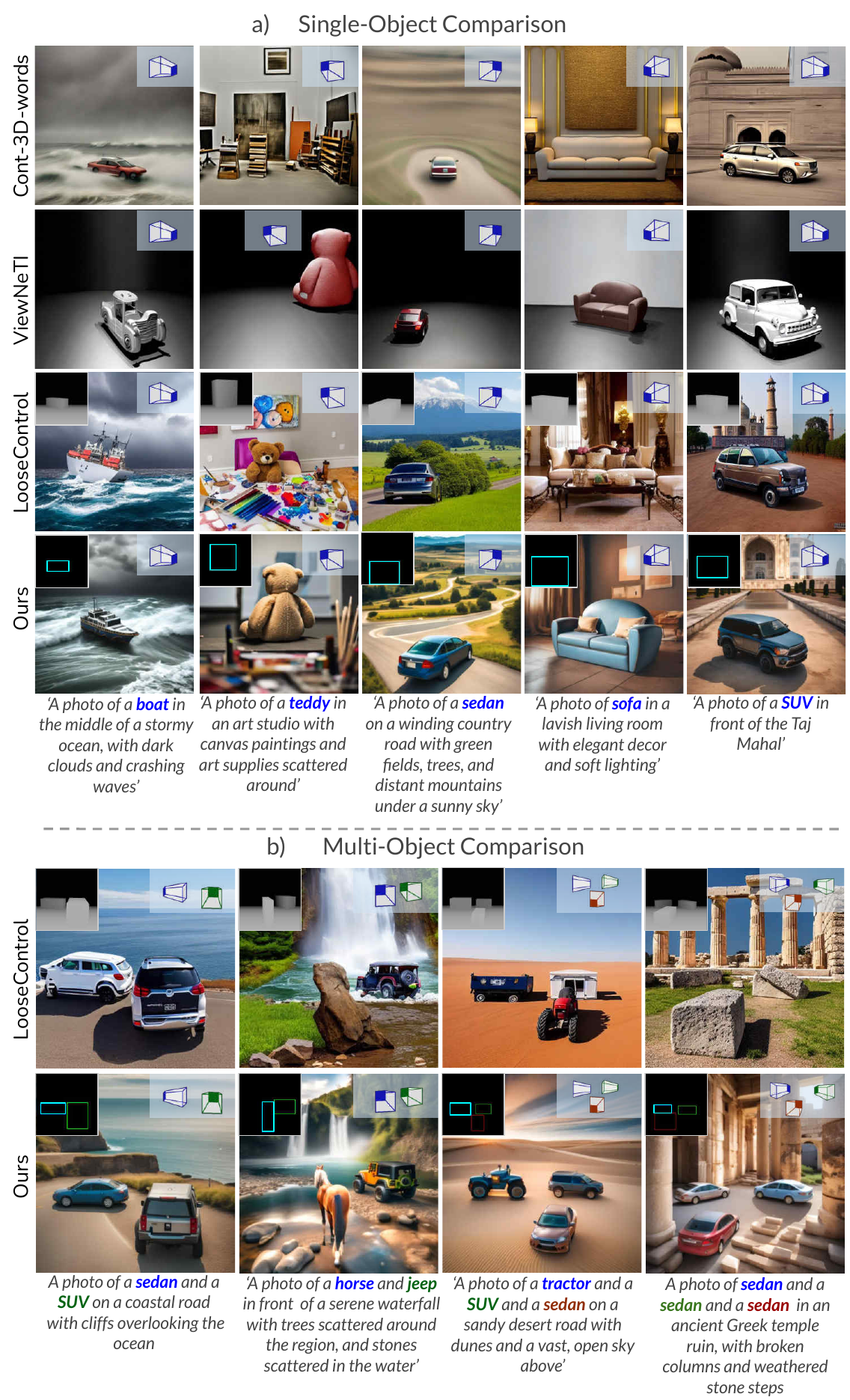}
    \vspace{-6mm}
    \caption{\textbf{Qualitative Comparison.} We compare our method against three baselines. Cont-3D-words~\cite{cont-words} does not generate the intended object whereas View-NeTI~\cite{view-neti} generates objects in plain backgrounds. LooseControl~\cite{loosecontrol} generates realistic compositions but does not follow the input orientation well. In contrast, our method aligns with the input text prompt and follows the input orientation, while generating realistic scenes.}
    \label{fig:baseline-compare}
\end{figure}

\noindent 
\textbf{Baseline comparison.} We compare our method to all three baselines on single-object scenes and additionally include multi-object scene comparison with LooseControl in Fig.~\ref{fig:baseline-compare} and Tab.~\ref{tab:quant-compare}.
Cont-3D-words morphs the generated objects into a \textit{sedan} shape seen during training and generates washed-out backgrounds. This results in poor text alignment and a lower percentage of intended object generation. ViewNeTI can generate better object shapes in a given orientation; however, it overfits to the black backgrounds seen during training, leading to poor text alignment. This is primarily because ViewNeTI does not accept ControlNet augmentations in its original form as it is designed for novel view synthesis. LooseControl generates realistic single-subject scenes following the given text prompt. However, in some cases, the object orientation is not followed (e.g., \textit{sofa, teddy}). For a multi-object generation (Fig.~\ref{fig:baseline-compare}b)), LooseControl either misses the object during generation (\textit{horse} in the second column and cars in the last column) or distorts the object shapes. LooseControl distorts the object to a \textit{box} like appearance (columns $3$ and $4$) as it learns LoRA over the original depth-conditioned ControlNet, which follows the depth input closely, resulting in an inferior object generation score. Further, though we adjust for a $180$-degree flip in computing Angular error for LooseControl, our method achieves significantly lower Angular error, demonstrating a strong orientation adherence.

\begin{table}[t]
\vspace{-2mm}
\centering
\begin{adjustbox}{width=0.95\linewidth}
\begin{tabular}{@{}cccc@{}}
\toprule
Single object  & Text Align. $\downarrow$ & \% Obj. Generated $\uparrow$ & Angular Err. $\downarrow$ \\ \midrule 
ViewNeTI~\cite{view-neti}      & $22.12$          & 0.920  & $0.596$         \\
Cont-3D-words~\cite{cont-words} & $29.88$          & 0.732  & $0.509$         \\
LooseControl~\cite{loosecontrol}  & $31.60$          & 0.656  & $0.385$         \\
Ours          & $\mathbf{32.98}$          & $\mathbf{0.968}$  & $\mathbf{0.198}$     \\ 
\toprule
Multiple object  & Text Align. $\downarrow$ & \% Obj. Generated $\uparrow$ & Angular Err. $\downarrow$ \\ \midrule 
LooseControl~\cite{loosecontrol} & $31.73$     &   0.778    & $0.372$      \\
Ours         & $\mathbf{33.93}$     &   $\mathbf{0.964}$    & $\mathbf{0.215}$      \\ \bottomrule
\end{tabular}
\end{adjustbox}
\vspace{-2mm} 
\caption{\textbf{Quantiative comparison.} We compute Text Alignment, using CLIP, \% of correct subject generation and Angular error between the predicted and input orientations.} 
\label{tab:quant-compare}
\vspace{-4mm}
\end{table}

\subsection{User study}
\begin{figure}[t]
    \centering
    \includegraphics[width=\linewidth]{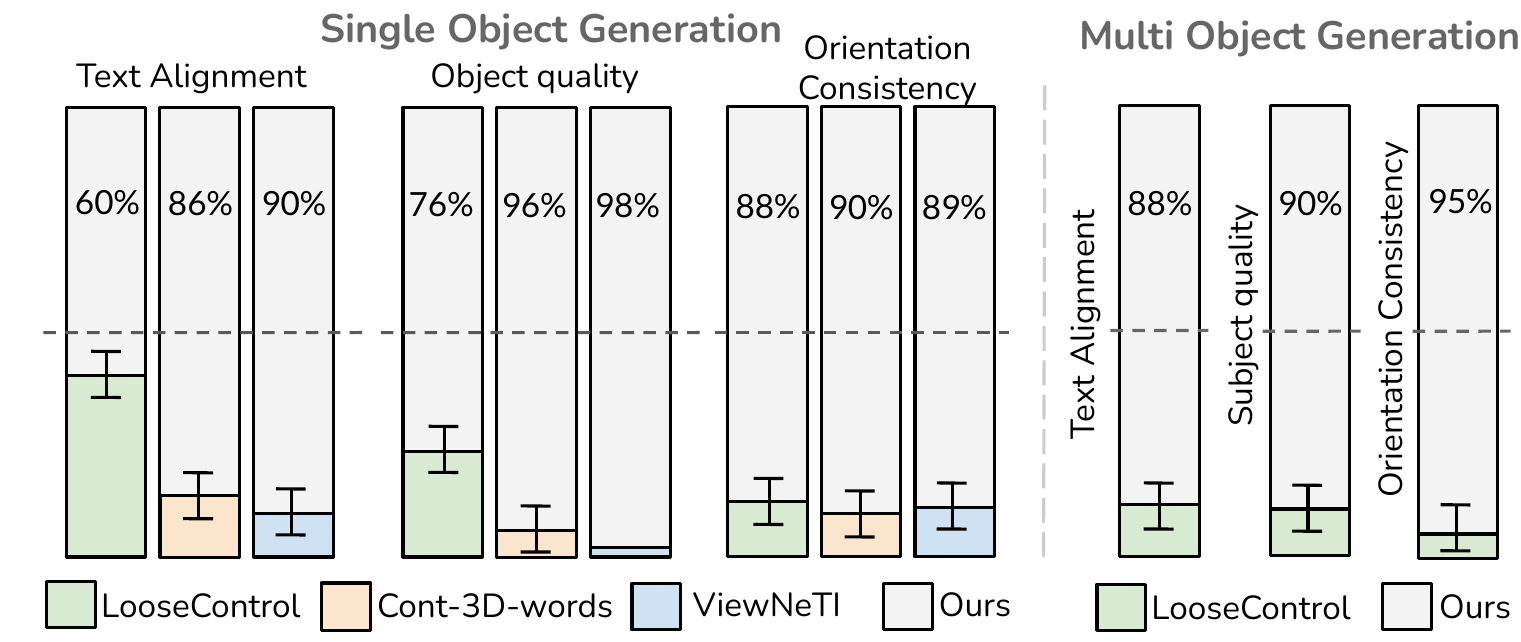}
    \vspace{-4mm}
    \caption{\textbf{User study.} We compare all methods on the three image metrics. Each bar indicates the fraction of people that preferred our result (gray) vs the baseline (other color).}
    \vspace{-4mm} 
    \label{fig:user-study}
\end{figure}

We conducted a user study with $57$ participants to compare all methods on \textit{text alignment}, \textit{object quality}, and \textit{orientation consistency}. Users rated $90$ image pairs for single-object scenes and $30$ for multi-object scenes, choosing the better image in each pair, sampled from our method and baselines. In total, we obtained $5130$ ratings for single-object and $1710$ for two-object scenes. Results (Fig.~\ref{fig:user-study}) show users preferred our method overall, with LooseControl scoring well in text alignment for single objects but falling short across all metrics in multi-object scenes.



\subsection{Personalization}
\begin{figure}[h]
    \centering
    \includegraphics[width=\linewidth]{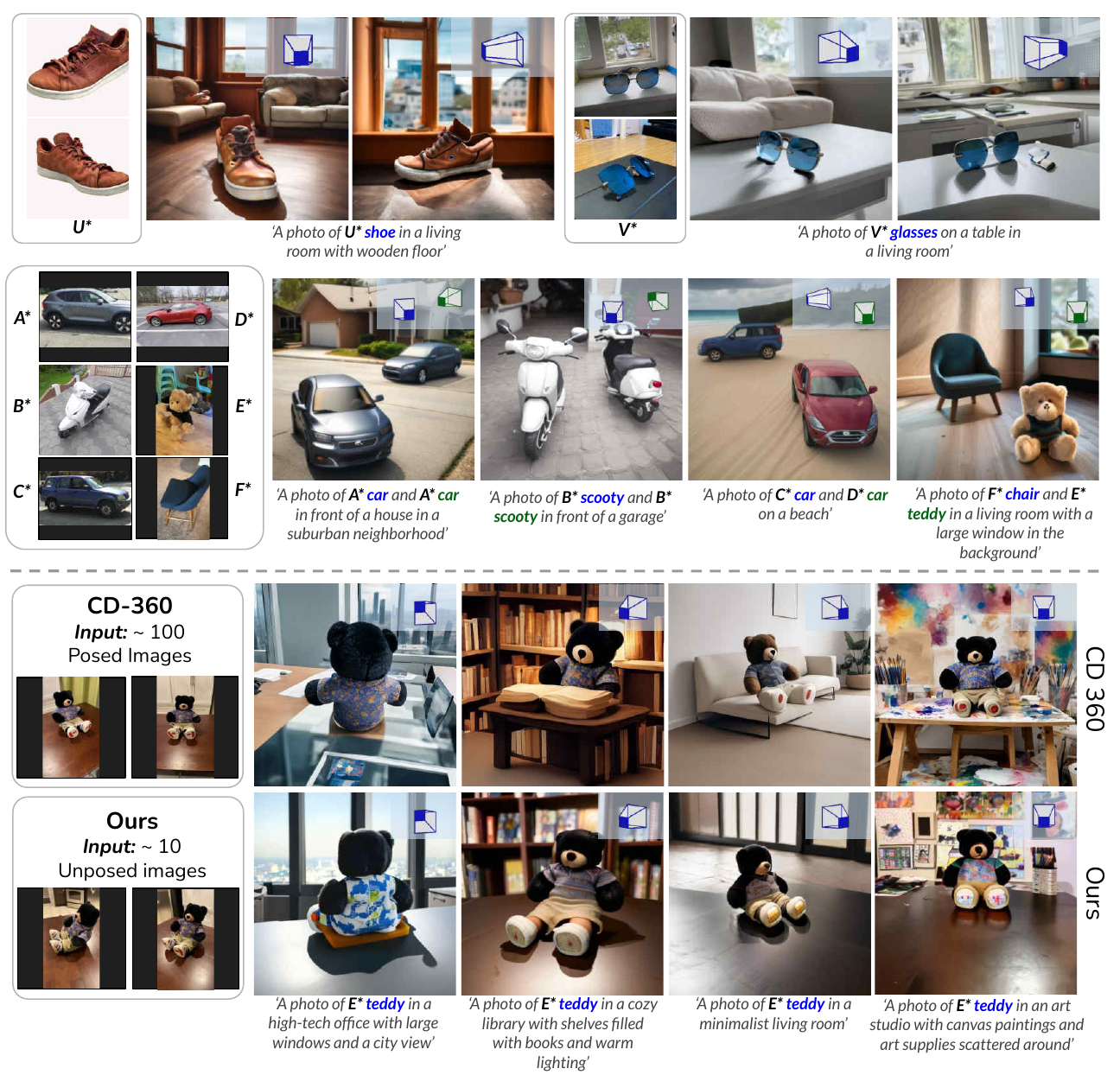}
    \vspace{-6mm}
    \caption{\textbf{Personalization.} Given a few ($\approx 10$) unposed images of an object, our method can personalize the diffusion models and allow for orientation control of the new object. Notably, our method can also generate scenes with two personalized objects with precise orientation control. Additionally, we compare our method with CustomDiffusion-360~\cite{cd-360} that uses $\approx 100$ posed images.}
    \vspace{-4mm} 
    \label{fig:personalize}
\end{figure}

We present personalization results in Fig.~\ref{fig:personalize}. 
With only $10$ unposed images of an object, our method can generate the object with precise orientation control in various contexts. 
Furthermore, we can jointly optimize Dreambooth~\cite{ruiz2023dreambooth} LoRA weights for two objects (e.g., a \textit{chair} and a \textit{teddy bear}), enabling multi-object personalization with object-centric orientation control. We compare our method with 
CD-360~\cite{cd-360} and achieve comparable performance. Unlike CD-360~\cite{cd-360}, which requires $\approx 100$ object images with camera pose, we require only a few unposed images, making it more convenient and user-friendly.

\subsection{Ablations} 
\begin{figure}[h] 
    \centering
    \includegraphics[width=\linewidth]{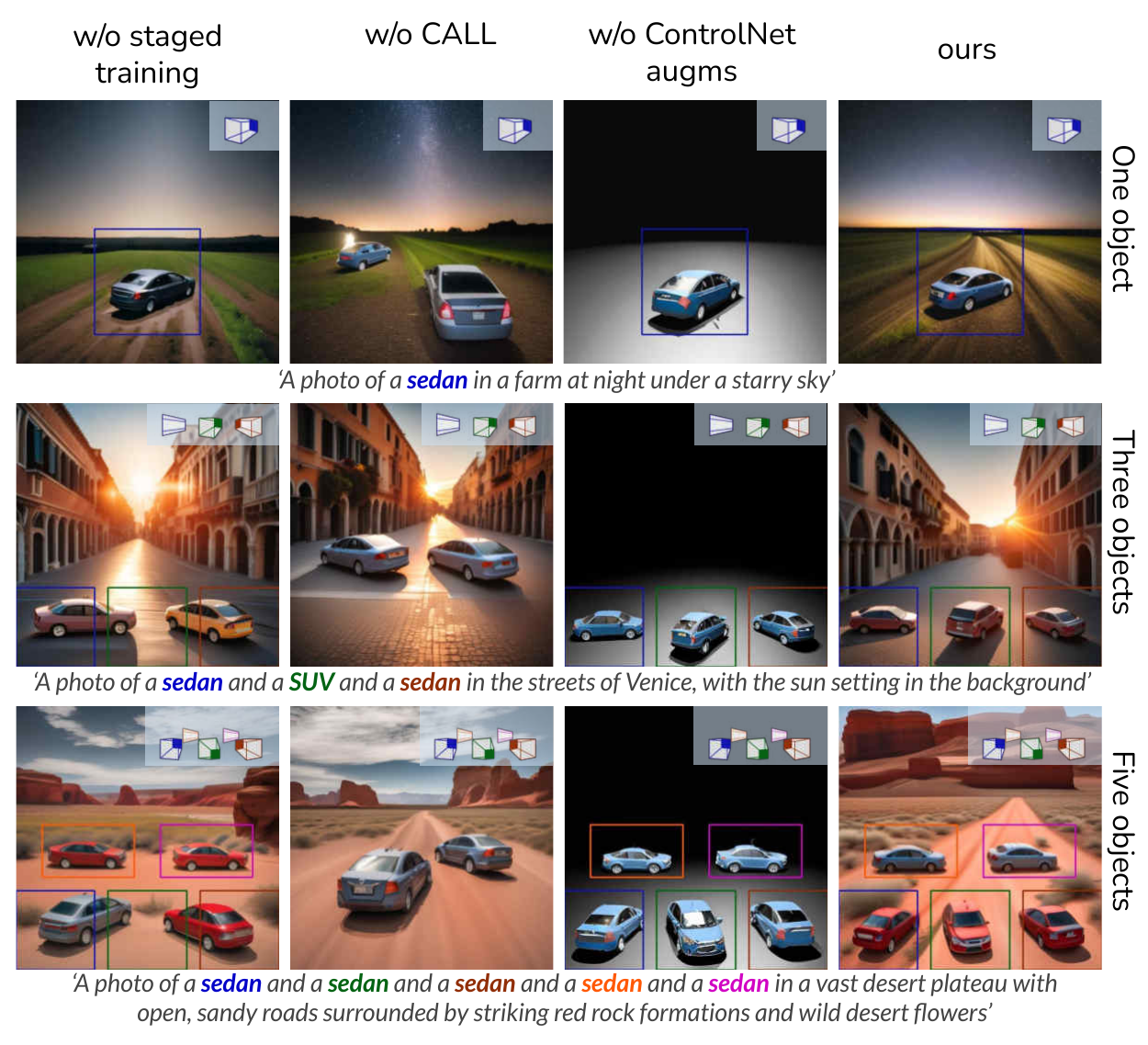}
    \vspace{-6mm}
    \caption{\textbf{Ablation studies.}
    We show the impact of several design
choices of our approach. Refer Sec.~\ref{subsec:ablations} for details.}
    \label{fig:ablation}
\end{figure}

\vspace{-1mm}
\label{subsec:ablations}
We present the results of the ablation study in Fig.~\ref{fig:ablation} on generated scenes with $1$, $3$, and $5$ objects. We focus on three key design choices and generate scenes with a variable number of objects:

\vspace{1mm}
\noindent 
\textbf{Staged training:} Training \textit{Compass Control} in a single stage results in poor adherence to the bounding boxes. This especially affects complex multi-object layouts, where some objects tend to \textit{leak} outside their box and suppress the generation of neighboring objects. This results in generating a lesser number of objects in the scene.

\vspace{1mm}
\noindent 
\textbf{CALL} is crucial for accurate orientation control. Without it the \textit{compass} token attends to irrelevant image regions, resulting in poor orientation control. Further, without CALL the object tokens entangle with each other during generation (a known issue in T2I models~\cite{attend-excite}). This results in generating a lesser number of objects in the scene. 

\vspace{1mm}
\noindent 
\textbf{Augmentations:} Without the ControlNet augmentations, the model overfits the training backgrounds, resulting in black backgrounds. Hence, ControlNet augmentation is necessary to generate objects in diverse contexts. 

%% file: sec/5_conclusion.tex
\vspace{-2mm}
\section{Conclusion and Discussion}
\label{subsec:conclusion}
\vspace{-1mm} 

\textbf{Limitations.} Our method struggles to control the orientation of objects that are occluded or have significant overlap.

\begin{wrapfigure}{r}{0.25\textwidth} 
    \centering
    \vspace{-3mm}
    \hspace{-3mm}
    \includegraphics[width=\linewidth]{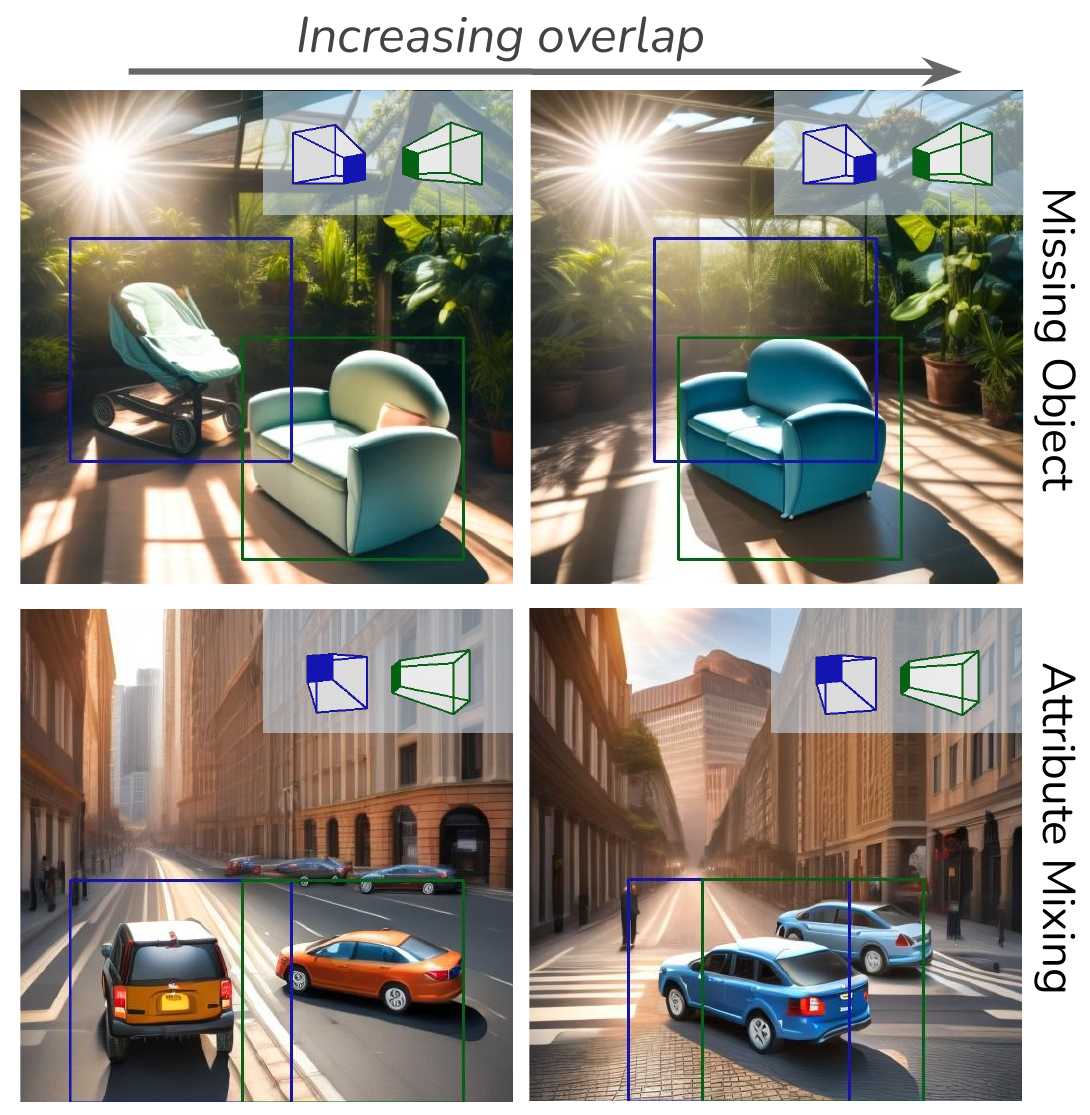} 
    \vspace{-4mm}
    \caption{Failure Cases}
    \vspace{-4mm}
    \label{fig:failure} 
\end{wrapfigure}

\noindent 
In these scenarios, the model either fails to generate one of the objects or mixes the attributes between objects (Fig.~\ref{fig:failure}). Further, our single-angle orientation representation is too simplistic for modeling complex non-rigid objects such as humans. In this work, we focus on presenting a generalized framework to condition text-to-image models on 3D controls that can be adapted easily for other representations.

\vspace{1mm}
\noindent 
\textbf{Conclusion.} In this work, we propose a method to condition pre-trained text-to-image models with 3D object orientation while preserving its rich image-generation capabilities. We train the conditioning module on a small synthetic dataset via staged training and involving attention regularization. These modifications enable strong generalization of the model, allowing for precise orientation control for unseen object categories and individual objects in a multi-object scene. Further, it can be seamlessly integrated with personalization methods to achieve orientation control of personalized objects. This work is a testament that text-to-image diffusion models innately have some form of 3D understanding, and interesting 3D controls can be obtained with appropriate conditionings.

\vspace{2mm}
\noindent 
\textbf{Acknowledgements.} We thank Srinjay Sarkar, Abhijnya Bhat, and Tejan Karmali for thoroughly reviewing the manuscript. This work is supported by Meesho and PMRF by the Government of India.

%% file: sec/X_suppl.tex
\clearpage
\setcounter{page}{1}
\maketitlesupplementary
\appendix 

\tableofcontents 

\vspace{6mm} 

\section{Project page}
Check the \href{rishubhpar.github.io/compasscontrol.github.io}{\texttt{project page}} for interactive visualizations of 3D orientation control.

\section{Controlling 3D orientation}
The main text primarily focused on an orientation control for a single angle. However, our method is not limited to single orientation control, and we present an experiment for controlling all three orientation angles in a single model. Specifically, we updated the pose injection network to take $3$ orientation angles as input to predict the pose token. We trained the model on flying objects - airplanes and helicopters- as rotation along all three axes is plausible for these objects. Specifically, we used six 3D assets from the web for these categories and followed the procedure in sec. 3.1 (main paper) to render the dataset. We present results for controlling all the $3$ orientation angles in Fig.~\ref{fig:3d-rot-single-obj} and ~\ref{fig:3d-rot-multi-object}. In Fig.~\ref{fig:3d-rot-single-obj}, we present rotation along all three axes for a fighter jet aircraft in three separate rows. Observe that, our method can precisely control all the object orientations along all the three axis. In Fig.~\ref{fig:3d-rot-multi-object}, we show the generalization of our trained model in controlling the orientation of a variety of objects. Notably, our model is not trained on birds or rockets. Still, it can generate consistent orientation-conditioned scenes following the text prompts. Note, that the compass shown in the figure is just for visualization purposes (can have an error of a few degrees).

\section{Additional Control}
\begin{figure}[h] 
    \centering
    \includegraphics[width=1.0\linewidth]{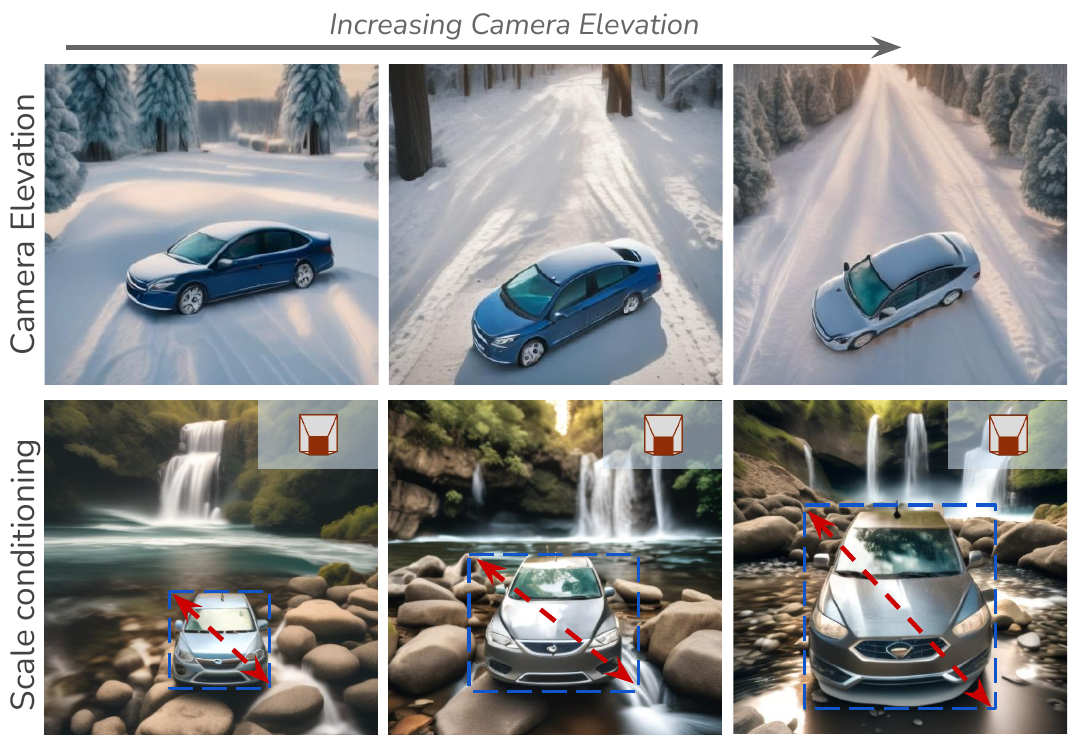}
    \caption{Additional Controls: (Top) Conditioning with camera elevation angle. (Bottom) Conditioning on object scale.}
    \label{fig:addn-control}
\end{figure}

\begin{figure*}
    \centering
    \includegraphics[width=0.90\linewidth]{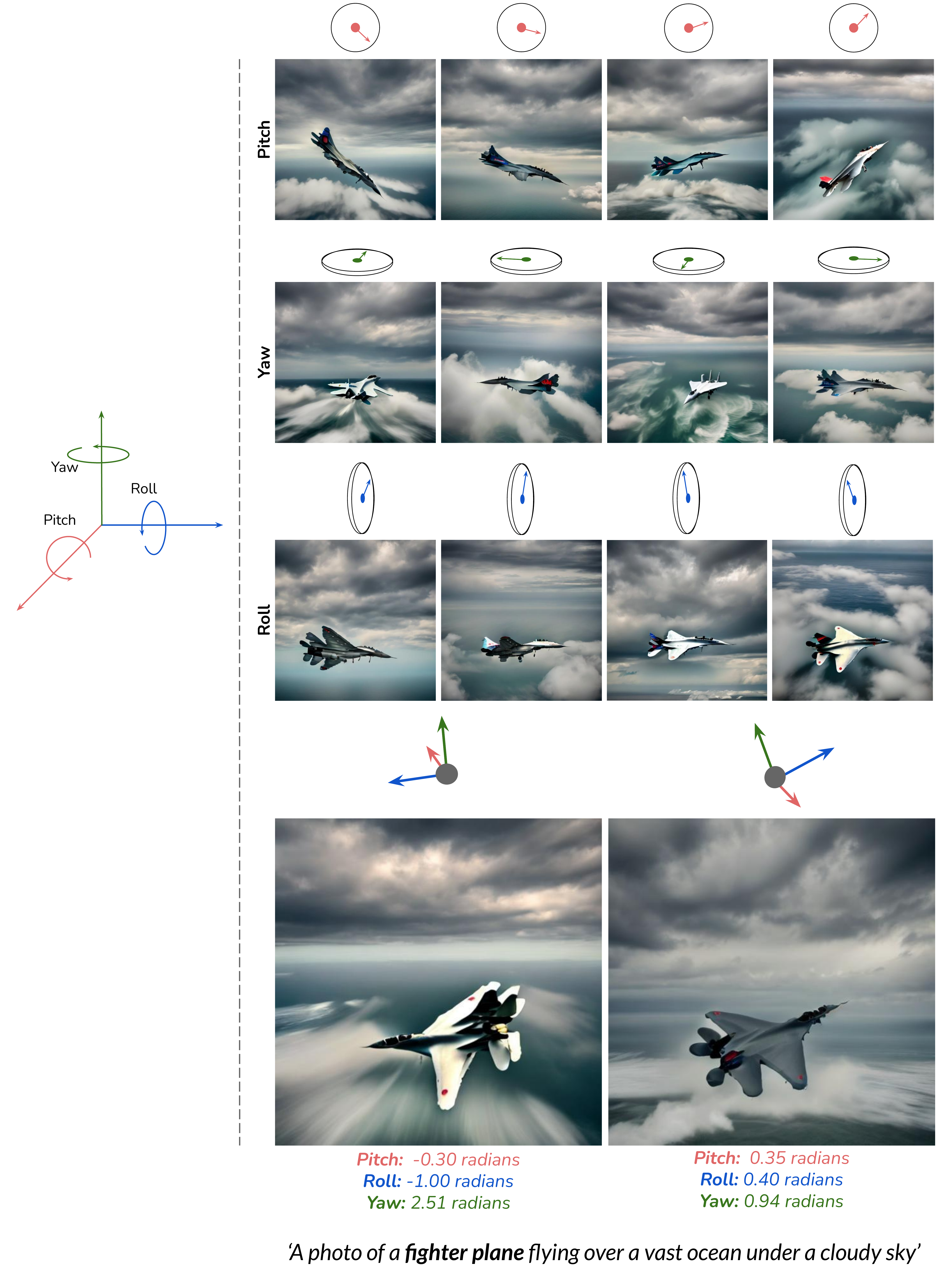}
    \caption{Conditioning on all three orientation angles for a single object.}
    \label{fig:3d-rot-single-obj}
\end{figure*}

\begin{figure*}
    \centering
    \includegraphics[width=0.95\linewidth]{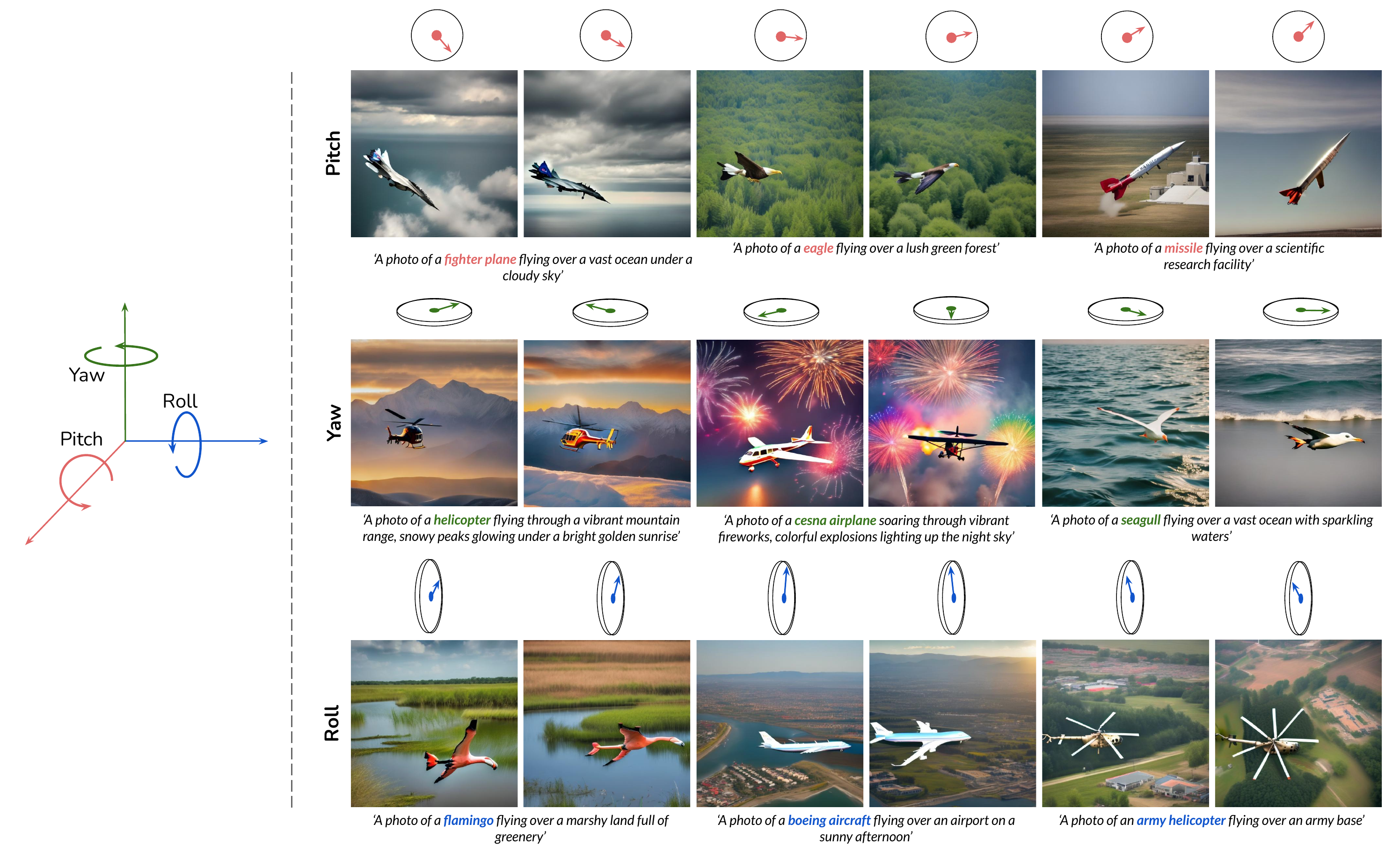}
    \caption{Conditioning on all three orientation angles}
    \label{fig:3d-rot-multi-object}
\end{figure*}

\textbf{Continuous control for camera elevation.}
Our proposed conditioning mechanism is generalized and can be adapted to achieve continuous camera elevation control in Fig.~\ref{fig:addn-control}. We generated a dataset with camera elevation variations and conditioned the denoising UNet on elevation angle. 

\noindent 
\textbf{Control for object scale.}
We can also precisely control the size of individual objects with additional conditioning on the object scale, as shown in Fig.~\ref{fig:addn-control}. Specifically, we condition the diffusion model with the length of the diagonal of a tight 2D bounding box.

\section{Generalization to StableDiffusion-XL}
\begin{figure}[h]
    \centering
    \includegraphics[width=1.0\linewidth]{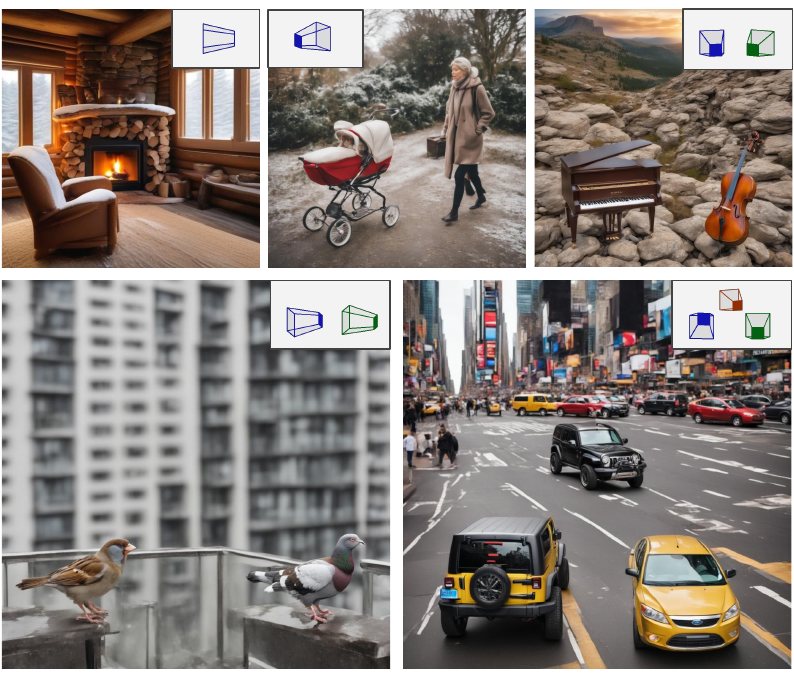}
    \caption{Compass Control on StableDiffusion-XL.}
    \label{fig:sd-xl-real}
\end{figure}

We have presented all the results on StableDiffusion-2.1 in the main paper. Our method also generalizes well to a larger StableDiffusion-XL backbone model shown in Fig.~\ref{fig:sd-xl-real}. The results demonstrate improved image quality with accurate orientation control of the generated objects.

\section{Diverse poses for non-rigid objects}
\begin{figure}[h] 
    \centering
    \includegraphics[width=1.0\linewidth]{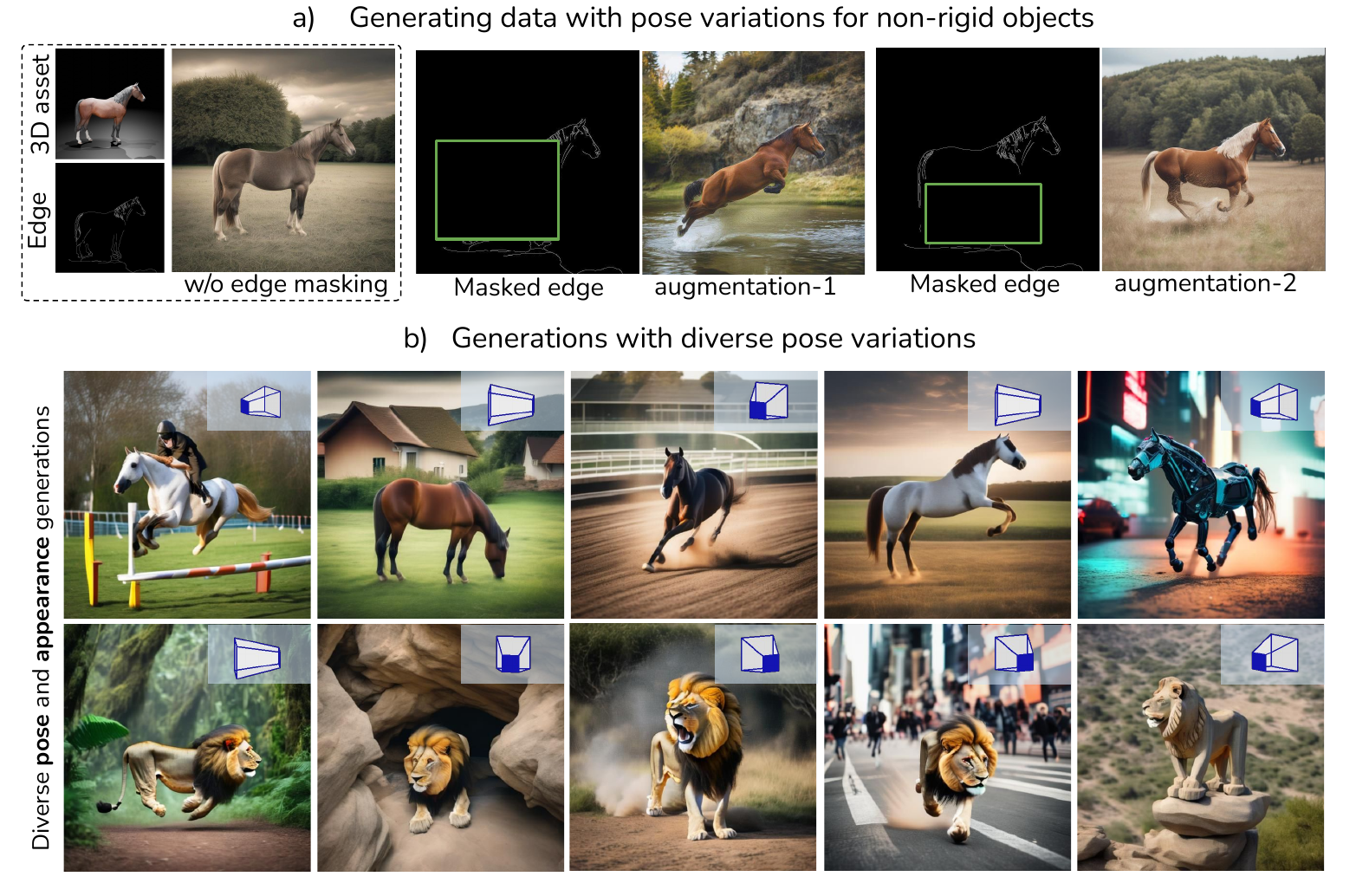}
    \caption{Pose variations for non-rigid objects}
    \label{fig:pose-variations}
\end{figure}

We build our dataset with only a few synthetic objects in their fixed canonical pose to generate our training data. This makes the model prone to overfitting on these poses for the non-rigid objects in the dataset - dog, horse, and lion. For instance, during inference, the model can generate only standing dogs in the given orientation. We generate augmentations with realistic pose variations in the training data to mitigate this. Specifically, we randomly mask some regions from the Canny Edge map and pass it to the ControlNet (Fig.~\ref{fig:pose-variations}a). This allows ControlNet to freely generate any plausible pose within the masked region. When trained with resulting augmentations, our method can generate diverse pose variations of non-rigid training objects while following the precise orientation as shown in Fig.~\ref{fig:pose-variations}b).

\section{Robustness to the 2D bounding boxes}
\begin{figure*}
    \centering
    \includegraphics[width=\linewidth]{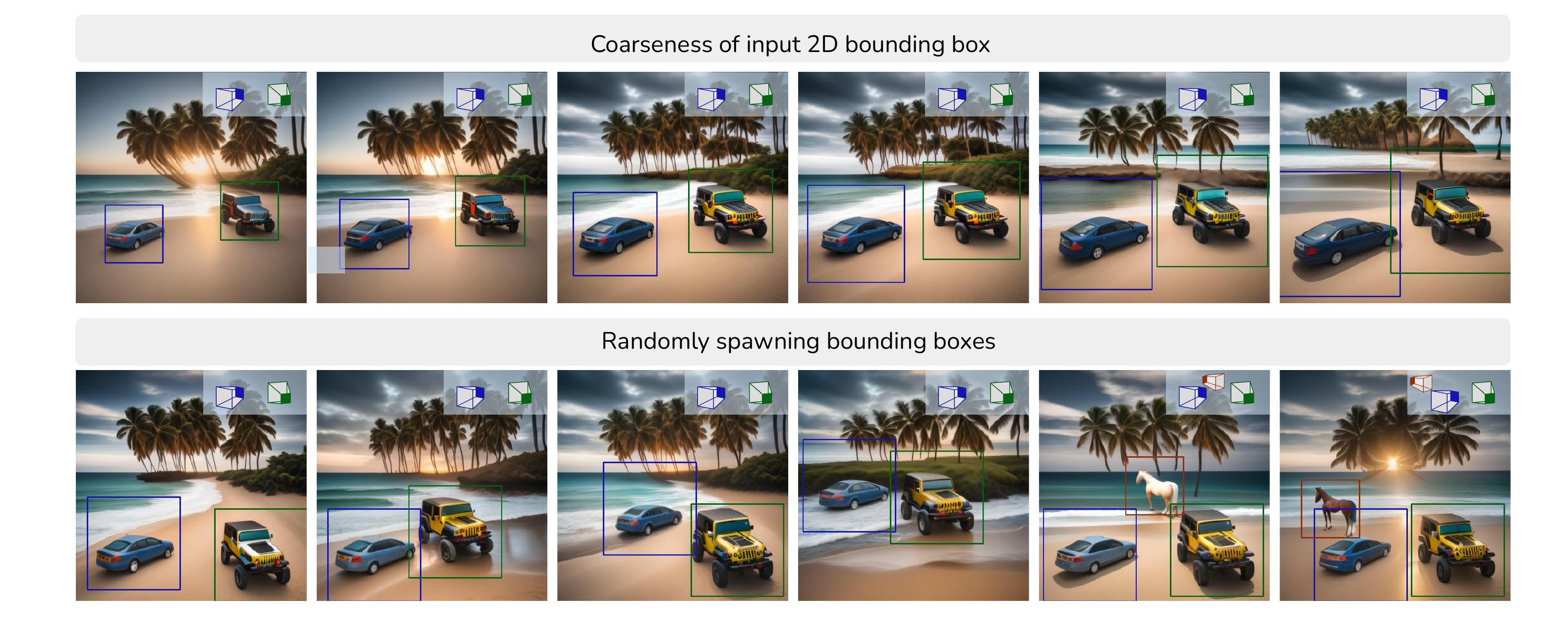}
    \vspace{-6mm}
    \caption{\textbf{Robustness of 2D bounding boxes.} Our method generates realistic scene compositions with different 2D bounding box sizes. Allowing for a loose bounding box during training provides this flexibility to the model to generate realistic scenes while coarsely following the input 2D box. Further, random non-overlapping boxes can also be spawned during inference without any degradation in quality. This robustness to the actual bounding box shape, reduces the burden on the user and is enabled by the \textit{loose} bounding box used during training.}
    \label{fig:bbox-user-input}
\end{figure*}

\textbf{Coarse bounding boxes.} We analyze the robustness of required 2D object bounding boxes during the inference. First, we analyze the effect of the coarseness of the bounding box on the generated scenes in Fig.~\ref{fig:bbox-user-input}. Our model does not generate objects that tightly occupy the provided bounding box. This is convenient for the user, as they don't have to provide an exact 2D bounding box. We present results for different bounding box sizes while keeping the center fixed. The model is robust to size changes and generates realistic scene compositions. The objects fall inside the box but they don't tightly fit the box. This provides more flexibility to the base generative model in generating more realistic scenes with relaxed constraints than conditioning on precise bounding boxes.

\vspace{1mm}
\noindent 
\textbf{Spawning random boxes.} In another experiment, we randomly spawn non-overlapping boxes, eliminating the user requirement to provide 2D boxes. The results are present in Fig.~\ref{fig:bbox-user-input}. Our method generates realistic compositions for these random layouts, with precise orientation control. The proposed design of using \textit{loose} bounding boxes during training, enables this, as the objects can adjust their size within the box region to make coherent scenes. 

\begin{figure*}
    \centering
    \includegraphics[width=0.85\linewidth]{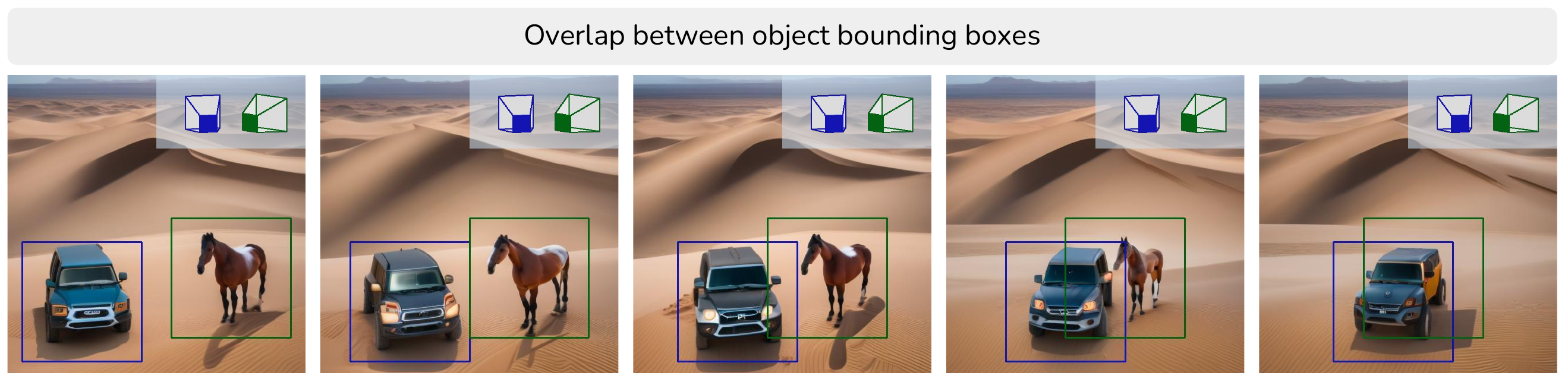}
    \caption{\textbf{Overlapping bounding boxes.} Our method can handle overlap between the two input bounding boxes up to a good extent. However, with a large overlap, the model struggles to generate accurate orientations (jeep in the fourth example), due to the mixing of pose tokens.}
    \label{fig:overlap-boxes}
\end{figure*}

\vspace{1mm}
\noindent 
\textbf{Overlapping boxes.} We present an ablation with the amount of overlap of 2D object boxes in Fig.~\ref{fig:overlap-boxes} during inference. Our method can handle the overlap between 2D boxes upto a good extent. On increasing the overlap the models' performance gracefully degrades in the pose control as the overlapping region is controlled by both the pose tokens (jeep in 4th example). With a large overlap in the bounding boxes, the model fails to generate both objects, and this is one of the limitations of our proposed approach, which is based on attention regularization. However, this limitation is common across all the bounding box conditioned or guided generative models. 

\section{Discussion with SoTA object-centric works.}
We compare the framework of our approach with recent works on object-centric 3D control in generation and editing with diffusion models. Particularly, we contrast our method with Neural Assets~\cite{neural-assets} and LooseControl~\cite{loosecontrol}, as these two are the closest method to ours. We present a comparison with both these methods at an approach level in Tab.~\ref{tab:compare-methods}.

\begin{table*}[h]
\begin{adjustbox}{width=\textwidth}
\begin{tabular}{ccccccc}
\hline
\multicolumn{1}{l}{} & Model type & Training data                                                                         & Input during inference                                                  & Novel categories & Input Representation                                                    & Personalization \\ \hline
LooseControl~\cite{loosecontrol}         & Generation & Real images (w 3D boxes)                                                              & 3D object boxes                                                         & Yes              & Explicit 3D (Depth)                                                     & No              \\ \hline
Neural Assets~\cite{neural-assets}        & Editing    & Real videos (w 3D boxes)                                                              & 3D object boxes                                                         & No               & \begin{tabular}[c]{@{}c@{}}Implicit\\ (List of bbox)\end{tabular}       & Yes              \\ \hline
Ours                 & Generation & \begin{tabular}[c]{@{}c@{}}Synthetic images\\ (w Orientation + 2D boxes)\end{tabular} & \begin{tabular}[c]{@{}c@{}}Orientation +\\ 2D object boxes\end{tabular} & Yes              & \begin{tabular}[c]{@{}c@{}}Implicit\\ List of orientations\end{tabular} & Yes             \\ \hline
\end{tabular}
\end{adjustbox}
\caption{Comparison with state-of-the-art approaches for object-centric control in the generation process.}
\label{tab:compare-methods}
\end{table*}

\section{Synthetic data generation} 
\begin{figure*}
    \centering
    \includegraphics[width=0.90\linewidth]{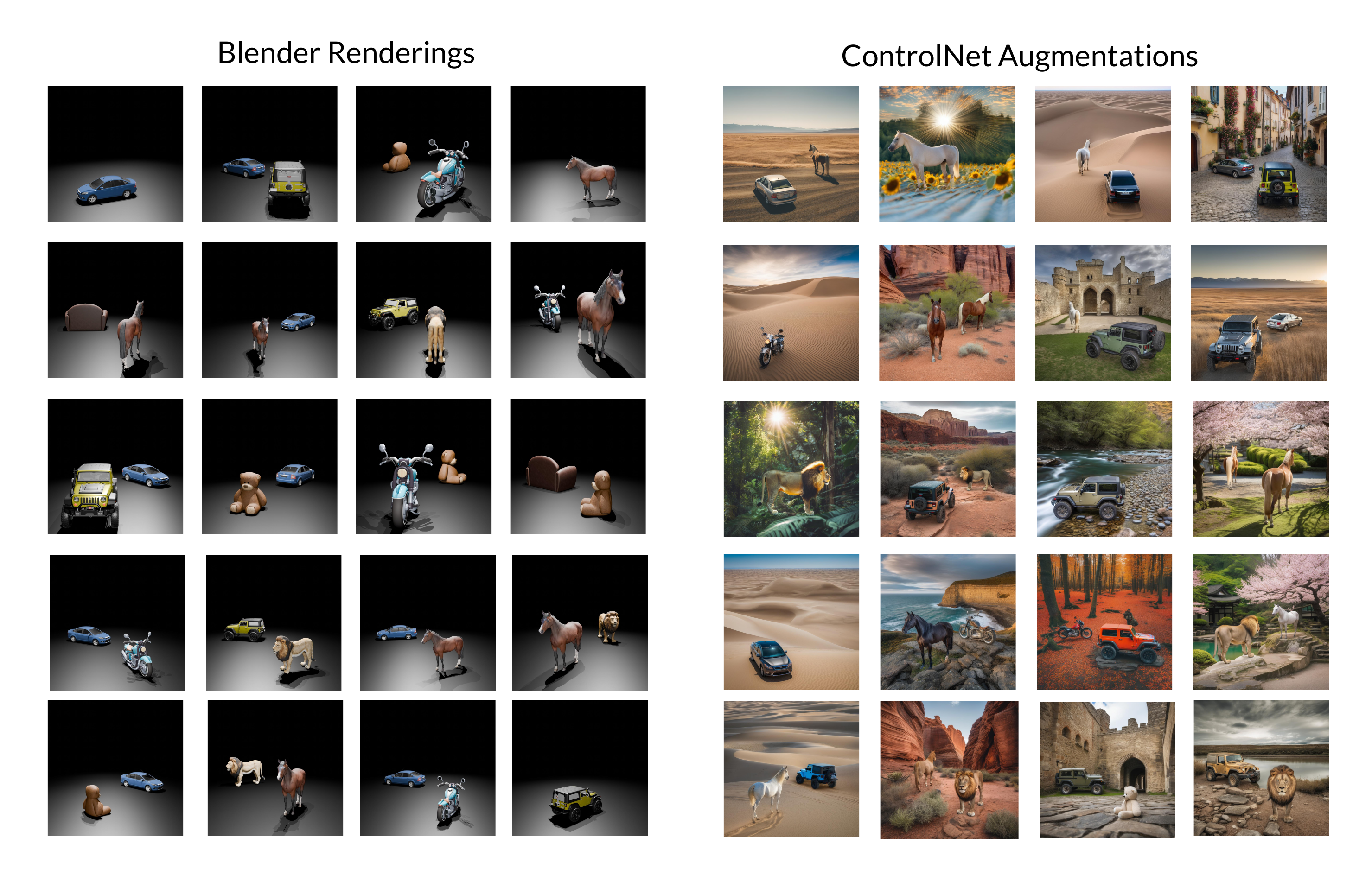}
    \caption{Samples from data generation process}
    \label{fig:synth-data-samples}
\end{figure*}

We render scenes with 3D assets in a Blender~\cite{blender} environment for our dataset. Specifically, we place an opaque floor on the $x-y$ plane and place a camera tilted slightly towards the ground at a fixed position. The scene is lighted using $3$ point lights of random intensity, placed at random locations. Once the environment is ready, we place the 3D assets at random locations and orientations and render the scene. For each rendered image we store the identity of the 3D assets in it, their respective orientations and 2D bounding boxes. We constrain the locations and orientations so that the object completely lies within the rendered image. Additionally, for two object scenes, we ensure that their 2D bounding boxes do not overlap. In all, we have $1000$ one-object scenes and 7900 two-object scenes. Some samples from the rendered images can be found in Fig.~\ref{fig:synth-data-samples}. 

\noindent 
However, training on this dataset alone leads to over-fitting to the plain backgrounds, as we have presented in the ablative experiments in the main text. Therefore, to generate the objects in diverse contexts, we augment the rendered scenes using Canny ControlNet~\cite{controlnet}. Specifically, given a rendered scene, we extract it's Canny map using OpenCV~\cite{opencv_library}, with the low and high thresholds set to $100$ and $200$ respectively. We use the following prompts for the augmentations:

\textcolor{teal}{\begin{enumerate}
    \item a photo of $\langle subject \rangle$ in a snowy forest, with a gentle snowfall and snow-covered trees
    \item a photo of $\langle subject \rangle$ in a vast desert with towering sand dunes and a clear blue sky
    \item a photo of $\langle subject \rangle$ in a medieval castle courtyard with ancient stone walls and archways
    \item a photo of $\langle subject \rangle$ in a sunflower field under a clear blue sky
    \item a photo of $\langle subject \rangle$ in a dense rainforest, with sunlight streaming through the canopy
    \item a photo of $\langle subject \rangle$ in a serene Japanese garden, surrounded by cherry blossoms
    \item a photo of $\langle subject \rangle$ on a rocky cliff overlooking a vast ocean
    \item a photo of $\langle subject \rangle$ by a riverside with wildflowers blooming nearby
    \item a photo of $\langle subject \rangle$ at a river's edge with stones scattered around
    \item a photo of $\langle subject \rangle$ in front of the Eiffel Tower at sunset
    \item a photo of $\langle subject \rangle$ in a vibrant autumn forest, with orange and red leaves carpeting the ground
    \item a photo of $\langle subject \rangle$ in a vast open plain, with golden grasses swaying in the wind and distant mountains on the horizon under a wide, clear sky
    \item a photo of $\langle subject \rangle$ on a cobblestone street in a quaint European village, with flower-filled balconies and historic buildings
    \item a photo of $\langle subject \rangle$ in a canyon with towering red rock formations, and scattered desert plants growing in the rocky terrain
\end{enumerate}}

We run this augmentation pipeline on all the rendered images, and do a manual filtering to remove the inconsistent generations. In all, we have $771$ single-object augmentations and $5239$ two-object augmentations.

\section{Orientation Regressor}
\begin{figure*}
    \centering
    \includegraphics[width=\linewidth]{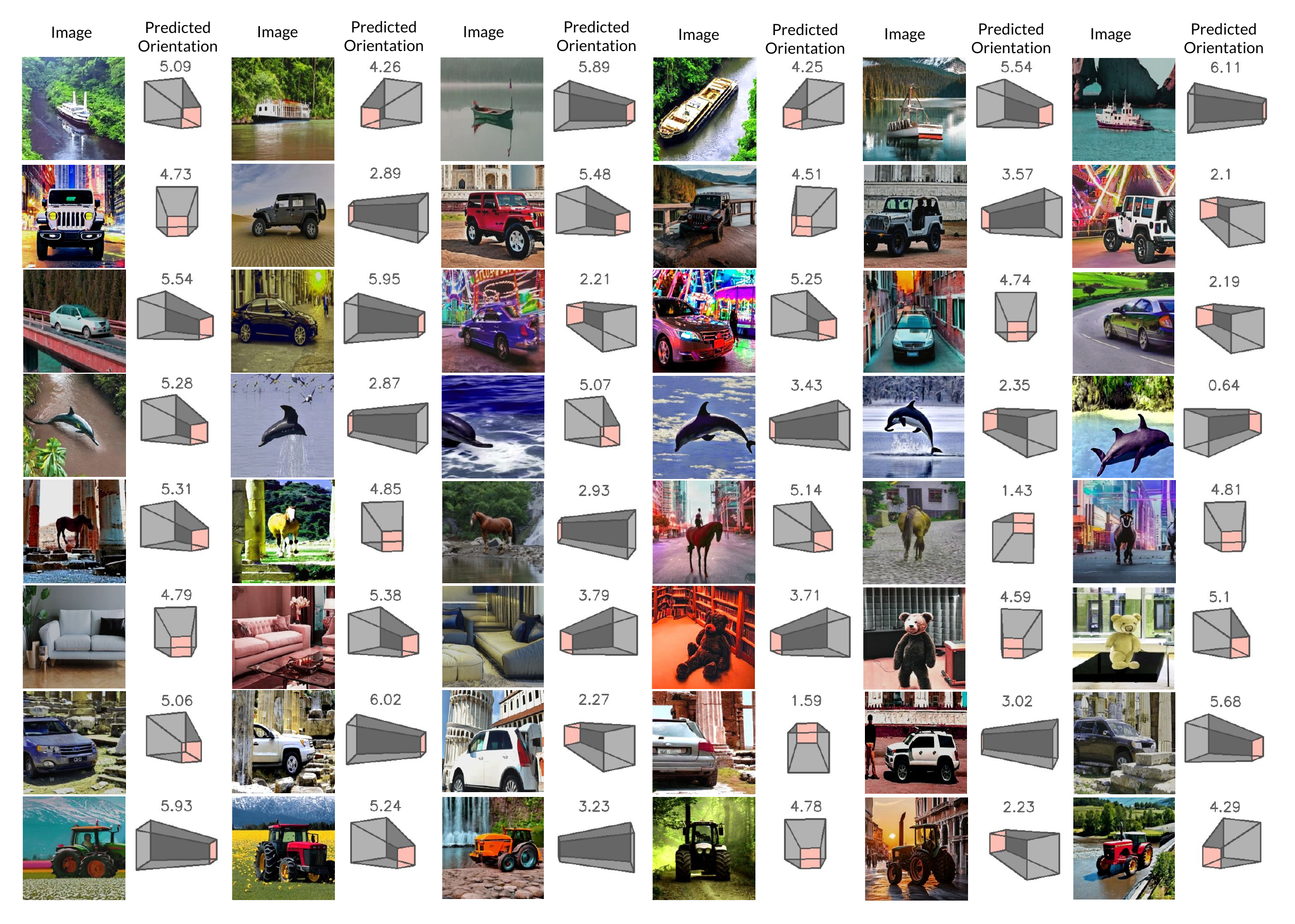}
    \caption{Predictions of the trained orientation regressor on unseen samples generated from Stable Diffusion~\cite{rombach2022high}. The model can predict the orientations accurately for the diverse unseen data and acts as a good critic to evaluate orientation consistency in generated images.}
    \label{fig:orient-regressor}
\end{figure*}

\noindent 
We train a neural network model to predict the orientation angle of an object in the generated image. We use a pretrained ResNet-18~\cite{he2016deep-resn} as the feature extractor and a mlp head consisting of two hidden layers of $128$ neurons, each with ReLU activations. Finally, we predict a single orientation angle $\theta$ along the up-axis (details in the main text - sec.3.1). We call this model \textit{orientation regressor} and train with a dataset of $35K$ images generated by rendering $30$ synthetic 3D assets of the test object categories followed by their canny ControlNet augmentations. This data is highly diverse, containing various backgrounds and object appearances, enabling the learning of an accurate orientation regressor. We train with a batch size of $128$, a learning rate of $5e-5$ for $95$ epochs with Adam optimizer. On an unseen test set of $8K$ images from the same distribution, the trained model achieves a mean angular error of $0.125$. Further, we present the results for evaluation on a completely unseen dataset, generated by Stable Diffusion~\cite{rombach2022high}, containing the test objects in Fig.~\ref{fig:orient-regressor}. We can observe that the trained orientation regressor predicts accurate orientations, and hence, it is a good estimator for evaluating pose consistency. In the case of multi-object scenes, we crop out the objects using Grounding DINO~\cite{liu2023grounding} and pass them to the \textit{orientation regressor}.

\section{Baseline details}
We provide implementation details for the baselines discussed in the paper. 

\subsection{ViewNeTI~\cite{view-neti}} ViewNeTI trains a small MLP to project the 3D camera pose to 3D view token. This token, along with the text prompt, is used to condition the text-to-image model. In the basic form, it is trained on a single scene with multi-view images and 3D camera poses. Once trained, the model can generate novel views for the trained scene. However, in an extended version, it is trained with multiple scenes to learn a generalizable view token. This token is then used for view control in text-to-image generation. For comparison, we use this version and train on our synthetic dataset of rendered multi-view scenes. Specifically, instead of conditioning on 3D camera pose, we condition on orientation angle $\theta$ and predict the view token. We train the model for $60K$ iterations on $1000$ multi-view images of $10$ assets. Note that because this model only supports a global view control, we train and evaluate it on only single object scenes for orientation control.

\subsection{Continous 3D Words~\cite{cont-words}} 
In this approach, a text-to-image diffusion model is conditioned on continuous 3D tokens to control 3D attributes such as lighting and object pose. They learn a generalizable \textit{3D word} in the text embedding space of the T2I model for each attribute, which is used along with the text prompt for conditioning. To learn the 3D word token, they use renderings of a single object and generate its augmentations with depth-conditioned ControlNet. However, it is essential that the 3D word token is disentangled from the object used for training. For this, they follow a staged training procedure: first learn the object's appearance (stage 1), and then learn the 3D attribute (stage 2). Following this, we train this model a single 3D asset, \textit{sedan}. We train for 5000 iterations in stage 1 and 15000 iterations in stage 2 (same as the original model). However, the trained model poorly generalizes to new objects as it is trained on a single object mesh (Fig.7 in the main text).

\begin{figure*}
    \centering
    \includegraphics[width=0.85\linewidth]{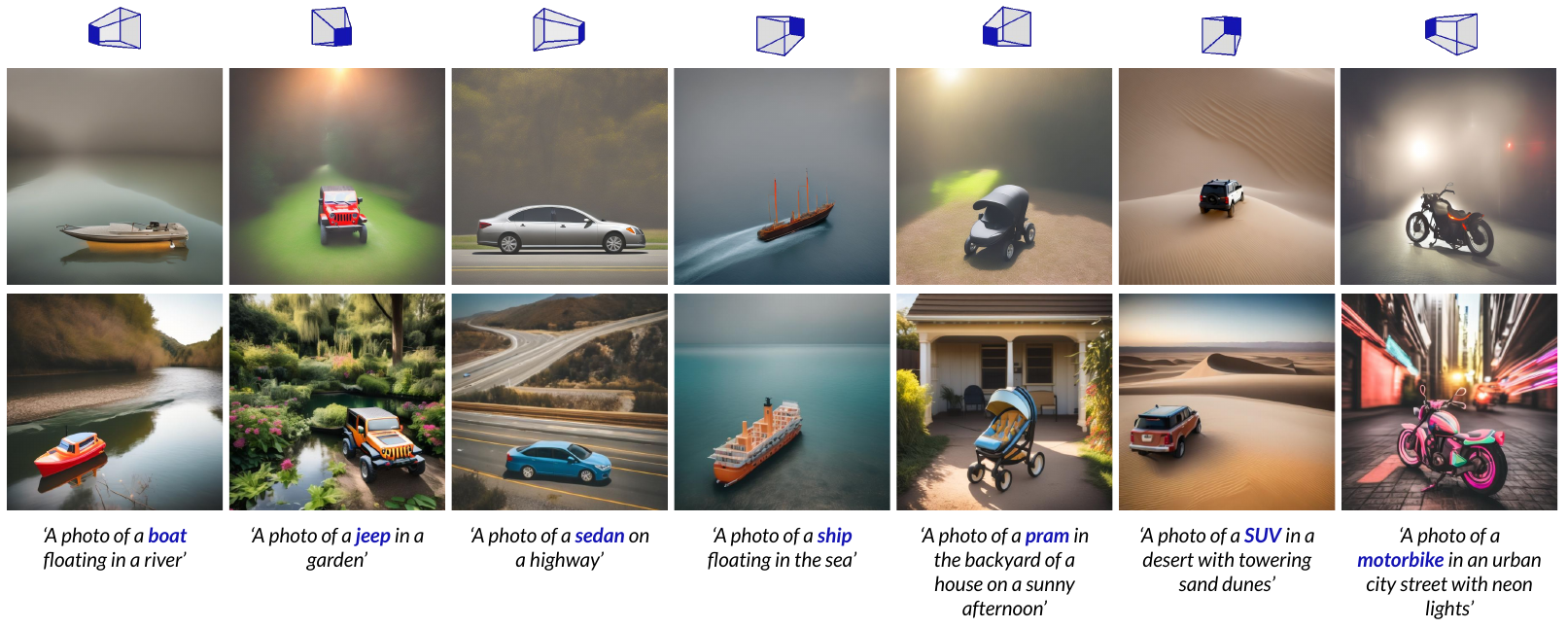}
    \caption{Comparsison with modified Continuous 3D Words~\cite{cont-words} trained on multiple assets. Compass control generates more realistic outputs and follows the text prompt better than the Cont-3D-Words trained on multiple object datasets.}
    \label{fig:mod-cont-3d-words}
\end{figure*} 

Here we present a variant of this model, which is trained on multiple 3D assets instead of just one (as proposed in their original paper). We use the same rendered images dataset as ours, and augment it using their proposed augmentation strategy. Notably, this dataset has diverse layouts and objects placed at random locations in the scene, making the learning process challenging. Since this model only allows for global control, we train and evaluate it on single object scenes only. We perform $30000$ training iterations in the first stage to learn the object appearance, followed by $70000$ iterations to learn the 3D word token. The comparison is presented in Fig.~\ref{fig:mod-cont-3d-words}. Our method achieves superior performance as compared to this baseline. The baseline struggles in pose control due to high diversity in the scene layouts, highlighting the importance of our attention localization mechanism CALL. Further, our backgrounds are much richer, as we use canny-conditioned ControlNet augmentations, which leads to richer augmentations. 

\subsection{LooseControl~\cite{loosecontrol}} 
LooseControl is a conditioning framework on text-to-image diffusion models that allows for 3D scene layout control. The framework is built on a depth-conditioned ControlNet model. However, instead of relying on accurate depth maps, which are often difficult to construct, LooseControl conditions the generation on coarse depth maps. Specifically, in this loose depth map, the scene boundaries are represented as planes, and the objects are represented by their loose 3D bounding boxes. LooseControl is implemented as a LoRA~\cite{hu2021lora} fine-tuning over depth-conditioned ControlNet model. This fine-tuning enables it to condition the generation using loose object depth maps also, against the accurate depth maps required by original ControlNet. In our experiments, we generate the loose depth maps by placing 3D bounding boxes in a Blender~\cite{blender} environment, and rendering the depth from camera viewpoint. Specifically, we randomly sample objects' locations and pose within the scene boundary and place a 3D bounding box for each object. Notably, one can control the object orientation by rotating the corresponding 3D bounding box in the input. We define a fixed template of 3D bounding box dimensions for each  test object in the dataset. The obtained depth maps are used to condition the model. We used the publicly available checkpoint for LooseControl in our evaluation. As this method allows for multi-object control, we compare both single and multi-object scenes. However, in experiments, we observe that LooseControl struggles to generate multi-object scenes with precise pose control and often resorts to generating bounding box artifacts. This is primarily due to the strong depth conditioning prior in the base depth ControlNet model, which is trained to follow exact depth maps.

\section{Additional Results}

\subsection{Comparisons}
We present additional baseline comparison results in Fig.~\ref{fig:supply-baseline-compare}. Our method follows the text prompts and generates objects following the input prompts

\begin{figure*}
    \centering
    \includegraphics[width=0.80\linewidth]{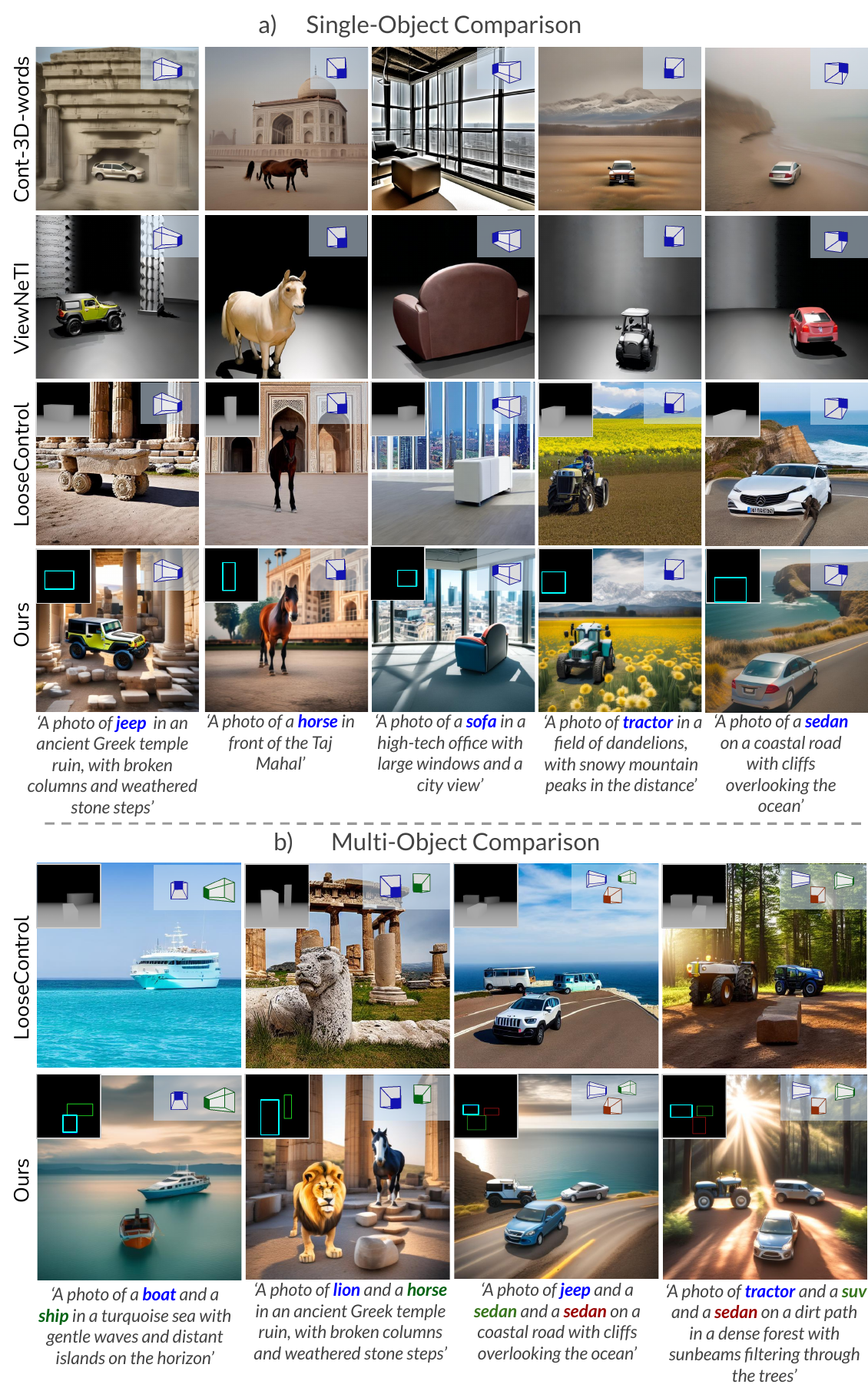}
    \caption{Additional comparison results with the baselines for single object and multi-object scenes.}
    \label{fig:supply-baseline-compare}
\end{figure*}

\begin{figure*}
    \centering
    \includegraphics[width=0.80\linewidth]{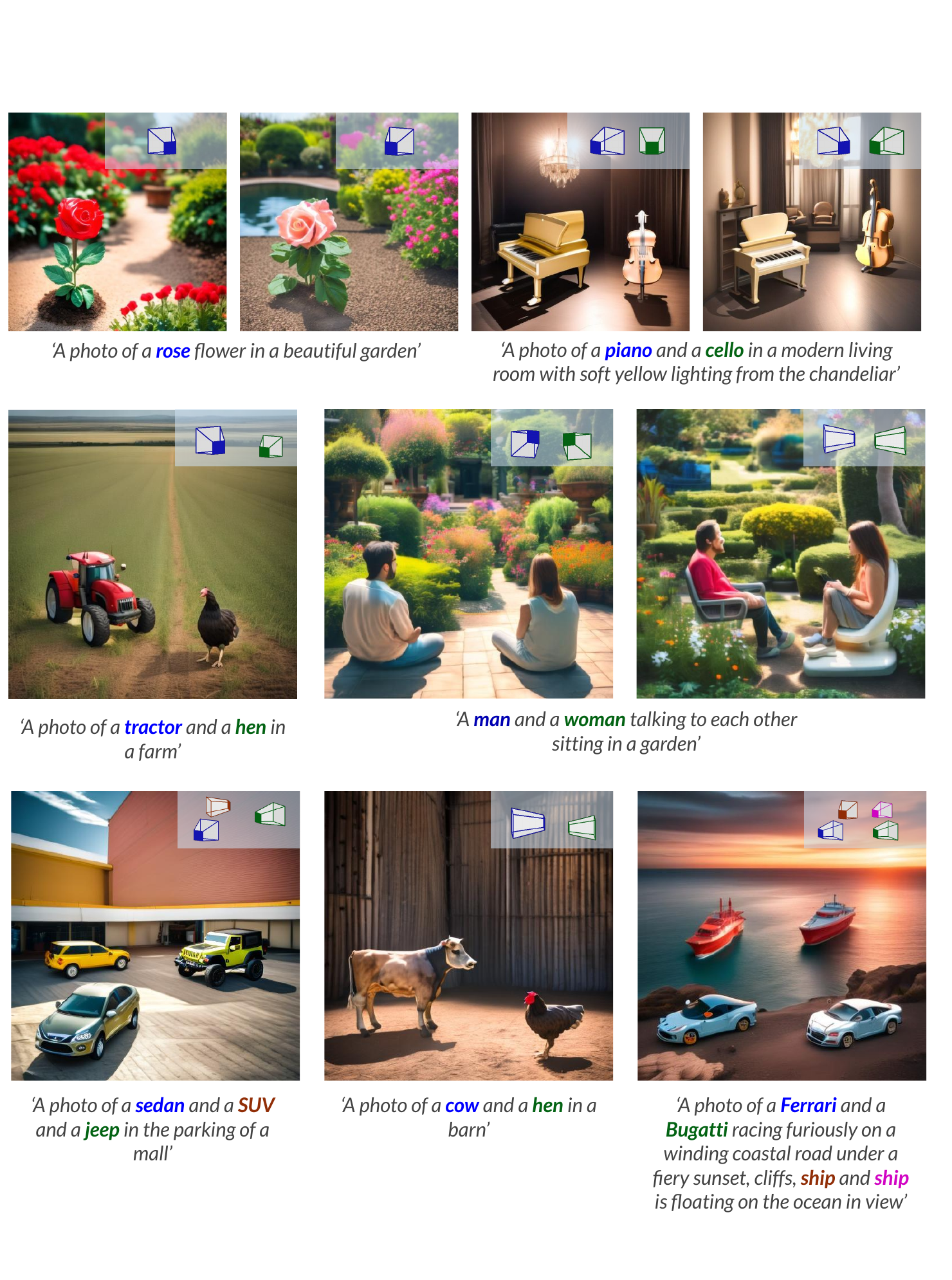}
    \caption{More qualitative results from our method, \textit{Compass Control}.}
    \label{fig:supply-baseline-compare}
\end{figure*} 

\section{Implementation Details}
\subsection{Method details}
We use Stable Diffusion v2.1~\cite{rombach2022high} as our base T2I model and use LoRA rank $4$ for fine-tuning its UNet. Our encoder model $\mathcal{P}$ is a lightweight MLP: three linear layers with ReLU. We train our model for $100K$ iterations with a batch size of $4$ with AdamW optimizer and a fixed learning rate of $10^{-4}$. We train first stage for $30K$ iterations with only single object scenes and the next stage for $70K$ iterations with mix of single and two subject scenes. We use SD-Xl for generating augmentations due to its higher realism. 

We keep the bounding box padding $\lambda=1.2$ for CALL. The training takes 24 hours on a single A6000 GPU, thus highly efficient.

\subsection{Evaluation dataset}
We randomly sample $10$ pose orientation in the range of ($0$,$360\deg$) for each prompt and object combination.
We used the following set of prompts for evaluation, containing single and two subject. In each prompt \textit{<subject>} is replaced with a single subject (e.g., jeep) or two subjects (e.g., jeep and sedan). Notably these prompts are different that the one used to generate ControlNet augmentations, to accurately evaluate model generalization. 

\textcolor{teal}{
\textbf{For road objects}
\begin{enumerate}
    \item A photo of $\langle subject \rangle$ in front of the Taj Mahal
    \item A photo of $\langle subject \rangle$ on the streets of Venice, with the sun setting in the background
    \item A photo of $\langle subject \rangle$ in front of the leaning tower of Pisa in Italy
    \item A photo of $\langle subject \rangle$ in a modern city street surrounded by towering skyscrapers and neon lights
    \item A photo of $\langle subject \rangle$ in an ancient Greek temple ruin, with broken columns and weathered stone steps
    \item A photo of $\langle subject \rangle$ in a field of dandelions, with snowy mountain peaks in the distance
    \item A photo of $\langle subject \rangle$ in a rustic village with cobblestone streets and small houses
    \item A photo of $\langle subject \rangle$ on a winding country road with green fields, trees, and distant mountains under a sunny sky
    \item A photo of $\langle subject \rangle$ in front of a serene waterfall with trees scattered around the region, and stones scattered in the water
    \item A photo of $\langle subject \rangle$ on a sandy desert road with dunes and a vast, open sky above
    \item A photo of $\langle subject \rangle$ on a bridge overlooking a river with mountains in the background
    \item A photo of $\langle subject \rangle$ on a dirt path in a dense forest with sunbeams filtering through the trees
    \item A photo of $\langle subject \rangle$ on a coastal road with cliffs overlooking the ocean
    \item A photo of $\langle subject \rangle$ in front of a historical castle with high stone walls and flags flying in the breeze
    \item A photo of $\langle subject \rangle$ in front of an amusement park with bright lights and ferris wheels in the background
\end{enumerate}}

\textcolor{teal}{\textbf{For water objects}
\begin{enumerate}
    \item A photo of $\langle subject \rangle$ on still waters under a cloudy sky, mountains visible in the distant horizon
    \item A photo of $\langle subject \rangle$ floating on a misty lake, surrounded by calm waters and serene, foggy atmosphere
    \item A photo of $\langle subject \rangle$ in the vast sea, with a clear blue sky and a few fluffy clouds
    \item A photo of $\langle subject \rangle$ in the middle of a stormy ocean, with dark clouds and crashing waves
    \item A photo of $\langle subject \rangle$ in a calm lake with lily pads and reeds growing near the shoreline
    \item A photo of $\langle subject \rangle$ on a river running through a dense jungle with vibrant green foliage
    \item A photo of $\langle subject \rangle$ in a mountain lake surrounded by pine trees and snow-capped peaks
    \item A photo of $\langle subject \rangle$ floating in a lagoon with tropical fish and coral visible beneath the water
    \item A photo of $\langle subject \rangle$ on a frozen lake with a snowy landscape surrounding it
    \item A photo of $\langle subject \rangle$ on a serene river at dusk, with reflections of the sunset on the water
    \item A photo of $\langle subject \rangle$ in the middle of a vast marshland with tall grasses and migratory birds flying overhead
    \item A photo of $\langle subject \rangle$ near a small waterfall cascading into a clear pool in a rocky area
    \item A photo of $\langle subject \rangle$ on a bay with large rock formations jutting out of the water
    \item A photo of $\langle subject \rangle$ in a turquoise sea with gentle waves and distant islands on the horizon
    \item A photo of $\langle subject \rangle$ in a narrow canal in an old European city, with historic buildings lining the waterway
\end{enumerate}}

\textcolor{teal}{\textbf{For indoor objects}
\begin{enumerate}
    \item A photo of $\langle subject \rangle$ in a modern living room setting with painted walls and glass windows
    \item A photo of $\langle subject \rangle$ in a minimalist living room
    \item A photo of $\langle subject \rangle$ in a cozy library with shelves filled with books and warm lighting
    \item A photo of $\langle subject \rangle$ in a high-tech office with large windows and a city view
    \item A photo of $\langle subject \rangle$ in an art studio with canvas paintings and art supplies scattered around
    \item A photo of $\langle subject \rangle$ in a rustic kitchen with wooden cabinets and a stone countertop
    \item A photo of $\langle subject \rangle$ in a lavish living room with elegant decor and soft lighting
    \item A photo of $\langle subject \rangle$ in a large dining hall with chandeliers and long tables
    \item A photo of $\langle subject \rangle$ in a traditional Japanese tatami room with sliding paper doors
    \item A photo of $\langle subject \rangle$ in a well-equipped gym with weights and fitness machines
    \item A photo of $\langle subject \rangle$ in a music studio with soundproof walls and musical instruments
    \item A photo of $\langle subject \rangle$ in a sunlit greenhouse filled with tropical plants
    \item A photo of $\langle subject \rangle$ in a children's playroom with colorful toys and posters on the walls
    \item A photo of $\langle subject \rangle$ in an underground wine cellar with wooden barrels and dim lighting
    \item A photo of $\langle subject \rangle$ in a cozy reading nook with a soft armchair and a small lamp
\end{enumerate}}